%% file: main.tex
\documentclass[lettersize,journal]{IEEEtran}
\usepackage{amsmath,amsfonts}
\usepackage{makecell}
\usepackage{array}
\usepackage[caption=false,font=normalsize,labelfont=sf,textfont=sf]{subfig}
\usepackage{textcomp}
\usepackage{stfloats}
\usepackage{url}
\usepackage{verbatim}
\usepackage{graphicx}
\usepackage{cite}
\usepackage{bbm}
\usepackage[dvipsnames]{xcolor}

\usepackage[whole]{bxcjkjatype} 
\usepackage[linesnumbered, ruled, vlined, commentsnumbered]{algorithm2e}
\SetCommentSty{mycommfont}
\usepackage{dsfont}
\usepackage{tikz}
\usepackage{graphicx}
\usepackage{comment}
\graphicspath{{figs/}}
\usepackage{amsmath}
\usepackage{amssymb}
 
\usepackage{amsthm}
\usepackage{bm}
\usepackage{dsfont}
\usepackage{times}
\usepackage{color}
\usepackage{mathtools}
\usepackage{ifthen}

\SetKwRepeat{Do}{do}{while}

\definecolor{mypink}{rgb}{0.858, 0.188, 0.478}

\newcommand{\argmax}{\operatornamewithlimits{argmax}}
\newcommand{\argmin}{\operatornamewithlimits{argmin}}
\newcommand{\prob}[2]{\mathbb{P}_{#1}\left[ #2 \right]}

\newcommand{\maxit}{c_{\mathrm{max}}}
\newcommand{\armservers}{Arm-Servers\textsuperscript{\textdagger} }
\newcommand{\armserver}{Arm-Server\textsuperscript{\textdagger} }
\newcommand{\desktop}{Desktop-PC\textsuperscript{\textdagger} }

\let\oldnl\nl
\newcommand{\nonl}{\renewcommand{\nl}{\let\nl\oldnl}}
\allowdisplaybreaks

\usepackage[firstpage]{draftwatermark}

\SetWatermarkText{\parbox{17cm}{\textcopyright 2025 IEEE. Personal use of this material is permitted.  Permission from IEEE must be obtained for all other uses, in any current or future media, including reprinting/republishing this material for advertising or promotional purposes, creating new collective works, for resale or redistribution to servers or lists, or reuse of any copyrighted component of this work in other works.}}
\SetWatermarkScale{1}
\SetWatermarkColor[gray]{0.4}
\SetWatermarkFontSize{0.3cm}
\SetWatermarkAngle{0}
\SetWatermarkHorCenter{4.1in}
\SetWatermarkVerCenter{2.48in}

\begin{document}

\title{CoverLib: Classifiers-equipped Experience Library by Iterative Problem Distribution Coverage Maximization for Domain-tuned Motion Planning}

\author{Hirokazu Ishida, Naoki Hiraoka, Kei Okada and Masayuki Inaba
\thanks{H. Ishida, N. Hiraoka, K. Okada and M. Inaba are with The University of Tokyo, 7-3-1 Hongo, Bunkyo-Ku, 113-8656 Tokyo, Japan (e-mail: \{h-ishida, hiraoka, k-okada, inaba\}@jsk.imi.i.u-tokyo.ac.jp). Corresponding author is Hirokazu Ishida.}}

\markboth{PREPRINT ACCEPTED BY IEEE Transactions on Robotics}%
{Ishida \MakeLowercase{\textit{et al.}}: A Sample Article Using IEEEtran.cls for IEEE Journals}


\maketitle

\begin{abstract}
Library-based methods are known to be very effective for fast motion planning by adapting an experience retrieved from a precomputed library. This article presents CoverLib, a principled approach for constructing and utilizing such a library. CoverLib iteratively adds an experience-classifier-pair to the library, where each classifier corresponds to an adaptable region of the experience within the problem space. This iterative process is an active procedure, as it selects the next experience based on its ability to effectively cover the uncovered region. During the query phase, these classifiers are utilized to select an experience that is expected to be adaptable for a given problem. Experimental results demonstrate that CoverLib effectively mitigates the trade-off between plannability and speed observed in global (e.g. sampling-based) and local (e.g. optimization-based) methods. As a result, it achieves both fast planning and high success rates over the problem domain. Moreover, due to its adaptation-algorithm-agnostic nature, CoverLib seamlessly integrates with various adaptation methods, including nonlinear programming-based and sampling-based algorithms.
\end{abstract}


\input{section/intro}
\input{section/formulation}
\input{section/method}
\input{section/setting}
\input{section/benchmark}
\input{section/discussion}
\input{section/conclusion}

\section*{Acknowledgments} We thank Yoshiki Obinata, who maintains the \armservers, for recommending their use. We also thank the reviewers for their valuable comments and suggestions.
 
\appendix[Details of Neural network archtecture, traning and inference] \label{sec:nn_appendix}
Neural network modeling, training, and inference are implemented using PyTorch framework \cite{paszke2019pytorch}.
The FCN used in vector modeling is composed of 5 linear layers (500, 100, 100, 100, 50), ReLU activation.
For vector-encoder modeling, FCN1 consists of 2 linear layers (50, 50), ReLU activation. The FCN2 is identical to the FCN used in vector modeling.
In the experiments above we used 2D CNN for Task 1 and Task 3, and 3D CNN for Task 4. The 2D CNN is composed of 4 convolutional layers with 8, 16, 32, and 64 channels, respectively. All convolutional layers have a padding of 1, kernel size of 3, and stride of 2 with ReLU activation. The output of the final layer is flattened and then fed into a final linear layer with 200 units and ReLU activation. The paired convolutional decoder network for pretraining the CNN encoder part has the following parameters: a linear layer with 1024 units, reshaped to $64\times 4\times 4$, followed by 4 inverse convolutional layers with 64, 32, 16, and 8 channels, and kernel sizes of 3, 4, 4, and 4, respectively. The stride is set to 2 and padding to 1 for all inverse convolutional layers. The 3D CNN encoder and pairing decoder are configured similarly to the 2D version.
All FCNs above have batch normalization layers before activation.

During training, we employ the Adam optimizer with a learning rate of 0.001. The batch size is set to 200. Early stopping is used with a \textit{patience} parameter of 10 (see Alg.~7.1 in \cite{goodfellow2016deep}). For training the classifier for Domain 4, we effectively used zstandard (zstd) compression/decompression to reduce RAM usage significantly, while keeping the entailed data processing overhead quite low.

To enhance the inference speed, we leverage PyTorch's JIT compiler. Additionally, we implement a batched inference strategy where $K$ FCNs are processed through tensorized operations, exploiting the design that all FCNs have identical architectures though the weights are different. Our batched implementation consolidates the weights of $K$ linear layers into a single tensor. This enhances the GPU-based parallelism and also reduces the overhead of memory transfer.

\bibliographystyle{IEEEtran}
\bibliography{main}

\newpage

\vspace{11pt}
\begin{IEEEbiography}[{\includegraphics[width=1in]{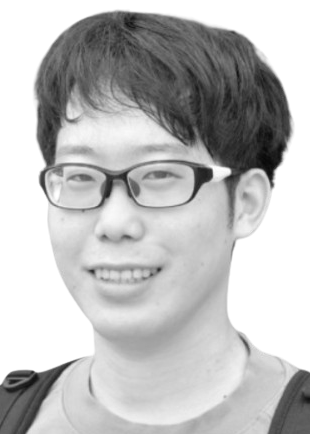}}]{Hirokazu Ishida}
received his BE in Aerospace Engineering from Osaka Prefecture University in 2016. He received his MS in Aerospace Engineering from the University of Tokyo in 2018. Since 2019, he has been a Ph.D. student in the Graduate School of Information Science and Technology at the University of Tokyo. His current research interests include the intersection of learning and planning in robotics.
\end{IEEEbiography}
\vspace{-5mm}

\begin{IEEEbiography}[{\includegraphics[width=1in]{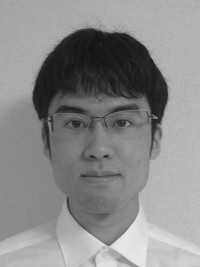}}]{Naoki Hiraoka}
received BE in Department of Mechano-Informatics from The University of Tokyo in 2019. He received MS in information science and technology from The University of Tokyo in 2021. From 2021, he is a doctoral student in the Graduate School of Information Science and Technology, The University of Tokyo and a Research Fellow of Japan Society for the Promotion of Science. His research interests include whole-body motion planning and control of humanoid robots.
\end{IEEEbiography}
\vspace{-5mm}

\begin{IEEEbiography}[{\includegraphics[width=1in]{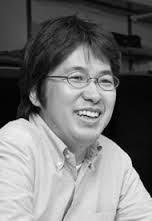}}]{Kei Okada}
received BE in Computer Science from Kyoto University in 1997. He received MS and PhD in Information Engineering from The University of Tokyo in 1999 and 2002, respectively. From 2002 to 2006, he joined the Professional Programme for Strategic Software Project in The University Tokyo. He was appointed as a lecturer in the Creative Informatics at the University of Tokyo in 2006, an associate professor and a professor in the Department of Mechano-Informatics in 2009 and 2018, respectively. His research interests include humanoids robots, real-time 3D computer vision, and recognition-action integrated system.
\end{IEEEbiography}
\vspace{-5mm}

\begin{IEEEbiography}[{\includegraphics[width=1in]{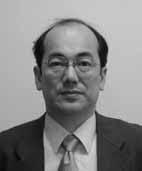}}]{Masayuki Inaba}
graduated from the Department of Mechanical Engineering, University of Tokyo, in 1981, and received MS and PhD degrees from the Graduate School of Information Engineering, University of Tokyo, in 1983 and 1986. He was appointed as a lecturer in the Department of Mechanical Engineering, University of Tokyo, in 1986, an associate professor, in 1989, and a professor in the Department of Mechano-Informatics, in 2000. His research interests include key technologies of robotic systems and software architectures to advance robotics research.
\end{IEEEbiography}

\vfill

\end{document}

%% file: section/intro.tex
\section{Introduction} \label{sec:introduction}
\IEEEPARstart{M}{otion} planning has been studied from two ends of the spectrum: global and local methods. Global methods, such as sampling-based motion planners (SBMP) like Probabilistic Roadmap (PRM) \cite{kavraki1996probabilistic} and Rapidly-exploring Random Tree (RRT) \cite{lavalle2001randomized}, are expected to find a solution if one exists, given enough time. However, these methods often require long and varying amount of computational time to obtain even a highly suboptimal solution and are sensitive to the complexity of the problem. On the other hand, local methods, including optimization-based approaches such as CHOMP \cite{zucker2013chomp} and TrajOpt \cite{schulman2014motion}, as well as SBMP with informed sampling, can quickly find a high-quality solution given a good initial guess, but struggle with problems that are far from the initial guess. There is a trade-off between plannability and speed when comparing global and local methods. In practice, it is often possible to define the scope of the \textit{domain}\footnotemark, which can be helpful in circumventing this trade-off and enabling quick solutions to a broad range of problems within a specific domain of interest. This is common in many practical robotics applications. For example, in automation for factories or warehouses, the environment and task types are fixed, but the specifics of each task may vary. Similarly, in home service robotics, although the tasks are diverse, the tasks that act as bottlenecks are often known in advance (e.g., reaching into a narrow container).

\begin{figure}[t]
\centering
\includegraphics[width=0.85\linewidth]{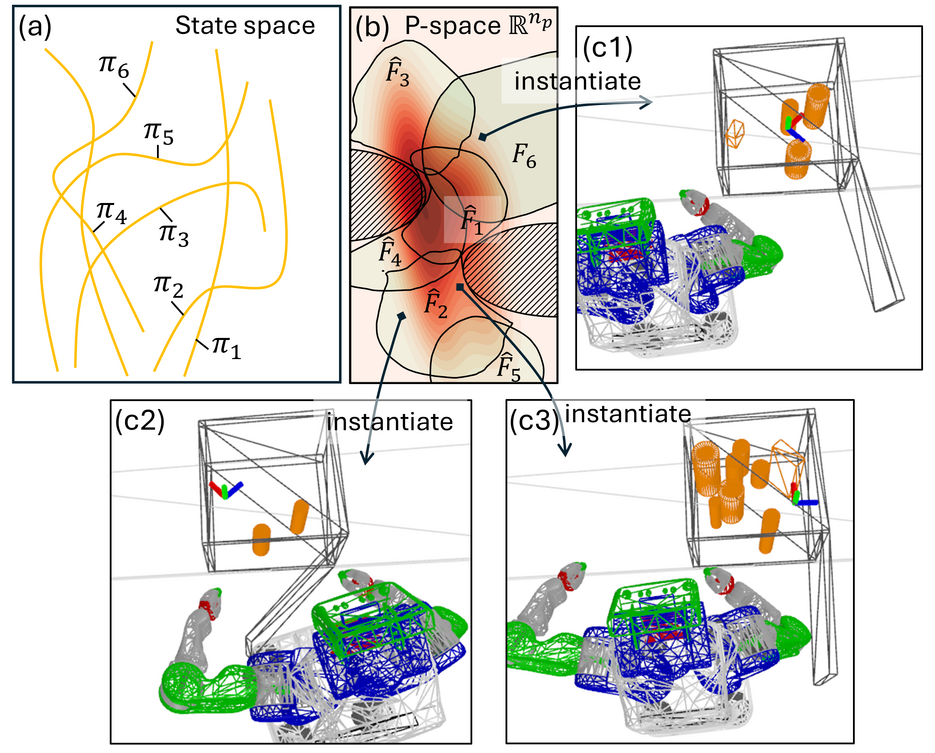}
\vspace{-2mm}
    \caption{Illustration of CoverLib's library  and the problem (P-) space. (a) The experiences (paths) $\pi_{1:6}$ defined in a state space, e.g. configuration space or phase space. (b) The problem distribution $p(\theta)$ (contour plot) in P-space covered by the union of the estimated adaptable regions $\hat{F}_{1:6}$ (green regions) of the experiences. Note that although depicted in 2D here, the P-space is often very high (e.g. dozens or hundreds) dimensional as it encodes problem settings including obstacle environments, constraints, success condition specifications, etc. (c) Instantiation of three problems from the P-space.}
\vspace{-4mm}
\label{fig:concept}
\end{figure}

\footnotetext{In this article, the \textbf{domain} refers to a pair of the distribution $p(\theta)$ of the problems and the (motion-planning) problem formulation.}
A promising approach to this end is to use a \textit{library} of \textit{experiences}\cite{jetchev2013fast, hauser2016learning, lembono2020memory, dantec2021whole, berenson2012robot, pairet2021path} reviewed in Section \ref{sec:nearest_neighbor}. An experience in this context refers to a previously solved solution path or trajectory, while the library is a collection of such experiences. The key insight is that if two problems are similar, the solution to one can be efficiently adapted to solve the other. 
Therefore, if the domain of interest is known a priori, the following preprocessing is beneficial: sample a problem, solve it, and store the solution as an experience, repeating this process multiple times.  Then, during the online phase, an \textit{adaptation}\footnote{In this article, \textbf{adaptation} refers to the process of solving a new problem by leveraging a previously solved experience.} algorithm can be used with a retrieved experience from the library to solve new problems quickly.

The library-based approach can be factorized into the \textit{high-level} and \textit{low-level} parts. The high-level parts are responsible for the construction and query of the library, while the low-level part is responsible for the internal adaptation algorithm. The high-level parts are adaptation-algorithm-agnostic (treating the adaptation algorithm as an input), allowing the library to be used with any adaptation algorithm, e.g., sampling-based or nonlinear programming- (NLP-) based. Although the low-level part has been extensively studied as reviewed in Section \ref{sec:nearest_neighbor}, the high-level counterpart has often received less attention in the community, with the exception of the line of seminal works \cite{islam2019provable, islam2021provably, islam2021alternative} by Islam et al. reviewed in Section \ref{sec:islams_work}. In fact, most library-based approaches consist of passive library construction and nearest neighbor search, neither of which reflects the domain structure or the characteristics of the adaptation algorithm.

The objective of this paper is to present CoverLib, a principled approach to precomputing libraries tailored to specific domains, which highlights the high-level parts of library-based methods.
We define coverage as the proportion of problems that can be solved using the library. CoverLib constructs a library with the explicit aim of maximizing this coverage.
CoverLib is designed to iteratively improve the coverage through an active procedure, where the next experience is actively determined to maximize coverage gain.
The key component of the algorithm is a set of binary classifiers that identify the \textit{adaptable regions} for each experience in the library.
An adaptable region, roughly speaking, is a region in the \textit{problem space} (P-space) where the experience is expected to be adaptable. By explicitly tracking the adaptable regions of the experiences added so far, we can actively sample an experience whose adaptable region covers the previously uncovered region effectively, i.e. that has high coverage gain. In the query phase, we utilize the classifiers to determine an experience expected to be adaptable for a given problem. Thus, CoverLib's library consists of pairs of experience and classifier, as illustrated in Fig.\;\ref{fig:concept}.

The active iteration in the library construction process makes CoverLib domain-tuned. This iteration is expected to concentrate experiences on the more challenging parts of the problem distribution and focus on dimensions with greater impact on adaptation, while ignoring those with less impact. Thus, it makes the library less susceptible to the curse of dimensionality. Furthermore, the query phase is also inherently domain-tuned, as the classifiers are trained to explicitly capture each experience's adaptability.

\section{Related work and Contributions}

\subsection{Library based motion planning} \label{sec:nearest_neighbor}
The adaptation algorithm, the low-level part, of library-based methods has been extensively studied. In NLP-based motion planning, the experienced solution serves as an initializer, and the optimization algorithm acts as the adaptation mechanism. This line of work has been explored using both unconstrained \cite{jetchev2013fast} and constrained \cite{hauser2016learning, lembono2020memory} formulations, and has even been applied to whole-body model predictive control (MPC) \cite{dantec2021whole} using differential dynamic programming. In SBMP, Lightning \cite{berenson2012robot} detects the parts of the experience that are in collision and \textit{repairs} such parts by generating a new path. Meanwhile, ERT(-Connect) \cite{pairet2021path} extracts a micro-experience from an experience retrieved from the library and deforms it to generate task-relevant node-transitions.

As for the high-level part, all of the above adopt nearest-neighbour search or some variants\footnote{Variants include combining nearest neighbour with multiple initialization methods \cite{lembono2020memory} and sorting the $k$-nearest neighbours to sequentially try them \cite{hauser2016learning}. } for the query part. Among them, \cite{hauser2016learning} is the pioneering work that argues for the sample efficiency of nearest neighbour search combined with random sampling-based library construction. The author argued that, under certain settings and with an adaptation strategy, an exponential number of experiences would be needed to achieve adaptation with a bounded optimality gap as the dimensionality of the P-space increases. This analysis relies on the fact, proven that there exists a small ball around each problem parameter, inside which the adaptation of the solution to the problem is guaranteed to be feasible with a bounded optimality gap. A similar conclusion could be drawn for the general nearest-neighbour based library method, with the key point being that the adoption of nearest neighbour search implicitly assumes each experience is homogeneously effective throughout the P-space with a constant distance metric.

However, such a ``small ball assumption" is often overly conservative, as suggested by the results and discussion in Sections \ref{sec:experiment} and \ref{sec:discussion}. In reality, adaptable regions are often quite large and irregularly shaped (see Fig.\;\ref{fig:reduction}), sometimes to the extent that a specific dimension is almost irrelevant for adaptability. This suggests that the adaptation algorithm has a certain nonlinear dimensionality reduction nature. The classifiers in Coverlib's library are expected to capture this irregularity and dimensionality reduction. Consequently, the number of experiences needed to cover the P-space is expected to be much smaller than the exponential number suggested by the small ball assumption.

\subsection{Constant-time motion planning by preprocessing} \label{sec:islams_work}
Islam et al. \cite{islam2019provable} proposed a library-based method that guarantees constant-time (or -iteration) solvability for a finite set of problems created by discretizing the continuous space. They presented a novel preprocessing method that iteratively precomputes an experience and its adaptable set (analogous to an adaptable region) pairs, saving them as a hash table. The iterative algorithm selects the new experience to solve at least one problem not yet covered by the union of the adaptable sets. This process continues until the feasible problem set is completely covered. This method avoids solving the entire problem set in a brute force manner and storing all the solutions. This approach is extended to the case where replanning is required to grasp a moving object \cite{islam2021provably}. Furthermore, a similar philosophy is found in their recently proposed library based method \cite{islam2021alternative} for a more specific problem setting.

However, these methods are based on discretization, which is inherently prone to the curse of dimensionality. In contrast, Coverlib's library is expected to be less prone to the curse of dimensionality, as stated in Section \ref{sec:nearest_neighbor}.

The adoption of a continuous P-space formulation (in Section \ref{sec:formulation}) does not strictly guarantee constant-time adaptability, which is a significant feature of Islam et al.'s methods. However, the proposed method offers a practical approach to this challenge by using Monte Carlo (MC) integration with a large number of samples to effectively validate the classifier's false-positive rate. To enable this MC integration without prohibitive computational costs, we developed a simple caching mechanism.

While our method shares similarities with Islam et al.'s methods \cite{islam2019provable, islam2021provably} in its iterative library population with awareness of uncovered problems, the continuous-discrete difference leads to three crucial algorithmic distinctions: (a) the use of a classifier, as the adaptable region is no longer enumerable; (b) enforcing the false positive constraint, as the classifier is never perfect; and (c) the active sampling of the next experience explicitly aiming to maximize coverage gain, as in practice only a limited number of iterations are allowed due to the high computational cost of obtaining the classifier.

\subsection{Trajectory prediction methods} \label{sec:trajectory_prediction}
While library-based methods implicitly realize problem-to-trajectory mappings via retrieval, explicit trajectory prediction methods aim to learn this mapping directly through regression models. Tang et al. \cite{tang2018learning} initially demonstrated success with a single neural network for trajectory prediction in agile quadrotor control. They later proposed a mixture-of-experts (MOE) model \cite{tang2019discontinuity} to better capture discontinuities in the problem-solution mapping. Building on this work, Merkt et al. \cite{merkt2021memory} introduced a novel trajectory clustering method using persistent homology, which is particularly effective for multimodal problem-solution mappings in obstacle environments. Their approach determines the number of clusters by analyzing $H_1$ groups (topological holes) in the persistent homology diagram, which is constructed using a distance matrix for trajectory segments. The determined number of clusters is then effectively used to set the number of experts in their MOE model.

The previous work excels in no-obstacle \cite{tang2018learning, tang2019discontinuity} or static obstacle \cite{merkt2021memory} environments, where the trajectory space structure closely reflects the domain structure and is exploited. Our work, however, aims for broader applicability in domains where such exploitation is infeasible, such as those with varying obstacle environment. Also, the key methodological distinction of CoverLib is its use of \textit{active} sampling and learning to explicitly maximize coverage. This feature sets it apart from trajectory prediction methods, which instead focus on minimizing prediction error through passive learning using a fixed training dataset.

\subsection{Other motion planning methods with learning} \label{sec:data_driven_general}
\textbf{Learning sampling distributions}: Learning sampling distributions is a promising approach to boost the performance of SBMP by focusing samples on relevant regions of the configuration space (C-space). Lehner et al. \cite{lehner2017repetition} learn the sampling distribution directly using Gaussian mixture models. Ichter et al. \cite{ichter2018learning} train a conditional variational autoencoder (CVAE) to generate not only domain-conditioned but also problem-conditioned samples. Chamzas et al. \cite{chamzas2019using, chamzas2021learning, chamzas2022learning} decompose the workspace into primitives and generate sampling distributions conditioned on these primitives, excelling at generating task-relevant sampling distributions for articulated robots' motion planning.

\textbf{Learning heuristic maps}: When planners have access to importance values (e.g., criticality or domain-relevance), such information can be harnessed in sample or connection generation for SBMPs. Zucker et al. \cite{zucker2008adaptive} learn a heuristic map of the workspace for fixed environments using reinforcement learning, which is then used for sampling in the C-space by solving inverse kinematics (IK). Terasawa et al. \cite{terasawa20203d} learn a mapping from occupancy grid maps to heuristic values in task space for Task-Space RRT. Ichter et al. \cite{ichter2020learned} propose a deep neural network to learn a mapping from configuration pairs and local geometric features to criticality, which excels for narrow passage problems.

\textbf{Learning latent spaces}: The key insight in this line of research is that there often exists a latent space that is far more simpler than the original C-space. MPNet \cite{qureshi2020motion} trains an encoder network modeling a mapping from a C-space to a latent space and a planning network that outputs the next configuration conditioned on obstacle information encoded in the latent space. Ando et al. \cite{ando2023learning} learn a mapping and its inverse between the C-space and a latent space, ensuring that straight lines in the latent space are collision-free. Ichter et al. \cite{ichter2019robot} take a factorized approach; they train distinct neural networks for forward/inverse-mapping to the latent space, dynamics in the latent space, and collision checking using the latent states.

\textbf{Learning roadmaps}: PRMs may also be considered as a data-driven approach because they precompute a graph in the C-space \cite{kavraki1996probabilistic}. Thunder \cite{coleman2015experience}, an extension of PRM, adopts a sparse data structure for memory efficiency and lazy collision checking for handling changing environments, outperforming the Lightning method \cite{berenson2012robot} in high-dimensional C-spaces.
Hierarchical dynamical roadmap (HDRM) \cite{yang2017hdrm} is an extension of PRMs that precomputes a correspondence between configuration and occupancy in the workspace. This precomputation enables to ignore implicitly defined edges that are in collision, resulting in massive speedup.

\subsection{Contribution statement} \label{sec:contribution}
Existing library-based motion planning methods, while promising for their speed, have been limited by low plannability due to their reliance on simplistic nearest neighbor methods in their high-level components. This paper addresses this limitation by proposing a principled approach to the high-level component for user-provided domains, enabling near global planner-level plannability while maintaining the high-speed planning capabilities of library-based methods.

Library-based approaches have demonstrated successful application across numerous motion planning settings, including inverse kinematics \cite{hauser2016learning}, NLP-based methods \cite{jetchev2013fast, lembono2020memory, dantec2021whole}, and SBMPs \cite{berenson2012robot, pairet2021path}. Their fundamental mechanism---comprising library building, retrieval, and adaptation---provides a versatile framework that can be adapted to various robotic motion planning scenarios. Furthermore, their core mechanisms are adaptation-algorithm-independent, allowing integration with other learning-based methods listed in \ref{sec:data_driven_general} through the low-level adaptation algorithm. Given this broad applicability and our method's ability to overcome the plannability limitation, our proposed approach has significant potential to advance multiple areas of robotic motion planning.

%% file: section/formulation.tex
\section{Problem formulation for library construction by coverage maximization} \label{sec:formulation}
Let $\theta \in \Theta$ be a (Problem-) P-parameter which uniquely represents a planning problem and $\Theta$ be a P-space. $\theta$ can be a vector, multi-dimensional array or other data structure. Note that we may refer to $\theta$ as a ``problem", although it is actually a vector representation of a problem. Let $p(\theta)$ be a distribution of the P-parameter. Let $p(c|\theta, \pi)$ be a distribution of the computational cost $c \in \mathbb{R}$ of an adaptation algorithm when solving a problem $\theta$ using an experience $\pi$. The computational cost is associated with a threshold $\maxit \in \mathbb{R}$, and adaptation is considered to fail if $c > \maxit$. Let $\hat{f}_{\pi}(\theta)$ be a real-valued estimator for $p(c|\theta, \pi)$. Using this, we define an estimated adaptable region \footnote{In the stochastic $c$ settings, there is no such thing as \textit{real} adaptable region, while it does exist in the deterministic $c$ setting.} as follows:
\begin{equation}
\hat{F}_{\pi}:=\{ \theta: \hat{f}_{\pi}(\theta) \leq \maxit \}
\end{equation}
where the region is associated with the experience $\pi$. It is important to note that the pair $(\hat{f}_{\pi}, \maxit)$ forms a binary classifier for $c \leq \maxit$.

Let $\pi_{1:k}$ be an indexed set of $k$ experiences. Also, for simplicity, let us write $\hat{f}_i := \hat{f}_{\pi_i}$ and $\hat{F}_i := \hat{F}_{\pi_i}$. The estimated coverage rate $r(\hat{f}_{1:k})$ is defined as follows:
\begin{equation}
r(\hat{f}_{1:k}) := \prob{\theta}{\theta \in \cup_{i=1}^{k} \hat{F}_{i}}
\end{equation}
where $\theta \sim p(\theta)$.
Let $i^*(\theta)$ be the optimal index selection function that selects the index of the experience minimizing the estimated computational cost for a given problem $\theta$:
\begin{equation} \label{eq:optimal_index}
i^*(\theta) := \argmin_{i \in \{1..k\}} \hat{f}_i(\theta).
\end{equation}
The false positive (FP) rate $\alpha(\pi_{1:k}, \hat{f}_{1:k})$ is defined as follows:
\begin{equation} \label{eq:fp_rate}
\alpha(\pi_{1:k}, \hat{f}_{1:k}) := \prob{(\theta, c)}{\theta \in \cup_{i=1}^{k} \hat{F}_{i} \land c > \maxit}/r(\hat{f}_{1:k})
\end{equation}
where $(\theta, c) \sim p(c|\theta, \pi_{i^*(\theta)}) \cdot p(\theta)$. This FP rate follows the standard definition in statistics, with the testing condition being the successful adaptation of $\pi_{i^*(\theta)}$ to $\theta$ in $c\leq \maxit$. The denominator in (\ref{eq:fp_rate}) represents the total probability of predicted successes (both true and false positives). The numerator represents the probability of false positive: when outcome is expected to be a success as $\theta \in \cup_{i=1}^{k} \hat{F}_{i}$ but is in reality not, as $c > \maxit$.

Now, the library construction problem is defined as the maximization of the estimated coverage rate while constraining the FP rate:
\begin{equation} \label{eq:main_problem}
\max_{\pi_{1:k}, \hat{f}_{1:k}} r( \hat{f}_{1:k}) \quad \mathrm{s.t.} \quad \alpha(\pi_{1:k}, \hat{f}_{1:k}) \leq \delta, k \leq K
\end{equation}
where $\delta \in \mathbb{R}$ is an FP rate threshold and $K \in \mathbb{N}$ is the maximum number of experiences in the library. In the above formulation, the user-tuning parameters at the formulation level are $\maxit$, $\delta$, and $K$. In practice, the value of $K$ might be determined by available computational resources or time constraints. By combining the output of (\ref{eq:main_problem}) and $\maxit$, we finally obtain the triplet $(\pi_{1:k}, \hat{f}_{1:k}, \maxit)$ as an experience library.

The obtained experience library is used for online planning as in Algorithm \ref{alg:online_planning}. As will be explained later, $\hat{f}_i$ is modeled by neural networks and evaluation of $\hat{f}_i(\theta)$ means a forward pass of the $i$-th networks to estimate the computational cost. Note that the $cautious$ parameter in Algorithm \ref{alg:online_planning} is explained in the next Subsection \ref{sec:comment}.

\begin{algorithm}
\caption{Planning using CoverLib}
\label{alg:online_planning}
    \nonl \textbf{procedure: } \textit{planWithCoverLib}($\theta |\pi_{1:k}, \hat{f}_{1:k}, \maxit$) \\
    \nonl \textbf{tuning param: $cautious = false$} (this paper):\\
    \small
    $i^* \leftarrow \argmin_{i \in \{1..k\}} \hat{f}_i(\theta)$ \tcp{Select the best (i.e. most relevant) experience index}
    \If{$cautious$ \textbf{and} $\hat{f}_{i^*}(\theta) > \maxit$}{
        \nonl \tcp{Don't attempt to solve if estimated cost exceeds threshold (i.e. $\theta$ is not covered)}
        \Return $nil$
    }
    solve $\theta$ by adapting experience $\pi_{i^*}$
\end{algorithm}

\subsection{Comment regarding the problem formulation} \label{sec:comment}
\textbf{Degree of freedom of $p(\theta)$ vs the dimension of $\Theta$:}
In this paper, the DOF $n_p$ of $p(\theta)$ represents the minimal number of parameter values needed to specify a sample in the distribution, which is distinct from the raw dimension of $\Theta$. We focus on $n_p$ rather than the dimension of $\Theta$ throughout this paper because $n_p$ remains invariant to representation changes.

To illustrate this concept, consider an environment where a fixed-shape sphere moves freely in 3D space. The sphere's position requires only three parameters ($x$, $y$, $z$ coordinates), setting the DOF of $p(\theta)$ at 3. Let us consider representing the P-parameter $\theta$ for this environment using a binary voxel map of shape $n_x \times n_y \times n_z$, for example. This representation creates a problem space $\Theta = \{0, 1\}^{n_x \times n_y \times n_z}$, where the dimension of $\Theta$ is $n_x \times n_y \times n_z$. This dimension grows cubically with resolution, even though the underlying system has only three degrees of freedom.

\textbf{Computation cost:} Computational cost $c$ is considered as a random variable in the formulation to account for the stochasticity of sampling-based adaptation algorithms. For deterministic adaptation algorithms, this formulation remains valid, as $c$ can be treated as a random variable with a Dirac delta distribution. 

\textbf{False positive rate constraint:} Due to the computational budget limitations for model training, non-negligible classification errors by $(\hat{f}_{i}, \maxit)$ are inevitable. These errors can be categorized into FP and false negative (FN) errors. FP errors occur when the classifier incorrectly predicts adaptation success, and vice versa for FN errors. While FP errors cause the classifier to be overconfident, FN errors make it more conservative. Although both lead to poor coverage, FP errors are more problematic for library construction. This is because once a region is incorrectly classified as covered due to an FP error, it will never be actively targeted for coverage again. In contrast, regions left uncovered due to the conservatism caused by FN errors could still actively be covered with additional experiences. Hence, it is crucial to adhere to the FP rate constraint $\alpha(\pi_{1:k}, \hat{f}_{1:k}) \leq \delta$ in (\ref{eq:main_problem}). On the other hand, the FN rate is indirectly considered in the optimization objective.

\textbf{In-advance feasibility detection in online phase:} We can easily infer problem feasibility in advance by $cautious=true$ in Algorithm \ref{alg:online_planning}. Note that verification of $[\hat{f}_{i^*}(\theta) > \maxit]$ in the algorithm is equivalent to verification of $[\theta \notin \cup_{i=1}^{k} \hat{F}_{i}]$, which means that the problem $\theta$ is not \textit{covered} by the library. This information can be took advantage of in higher level planning. For example, if a reaching action is deemed infeasible, the task planner can skip it, try another action, or request human assistance to reconfigure the environment. Additionally, task-and-motion-planning (TAMP) systems can utilize this information to accelerate their search process. However, as feasibility detection is not the focus of this paper, in the following experiments we always set $cautious=false$ in Algorithm \ref{alg:online_planning}, and proposed framework will proceed regardless of the feasibility inference result.

%% file: section/method.tex
\begin{figure*}[t]
    \centering
    \includegraphics[width=0.95\linewidth]{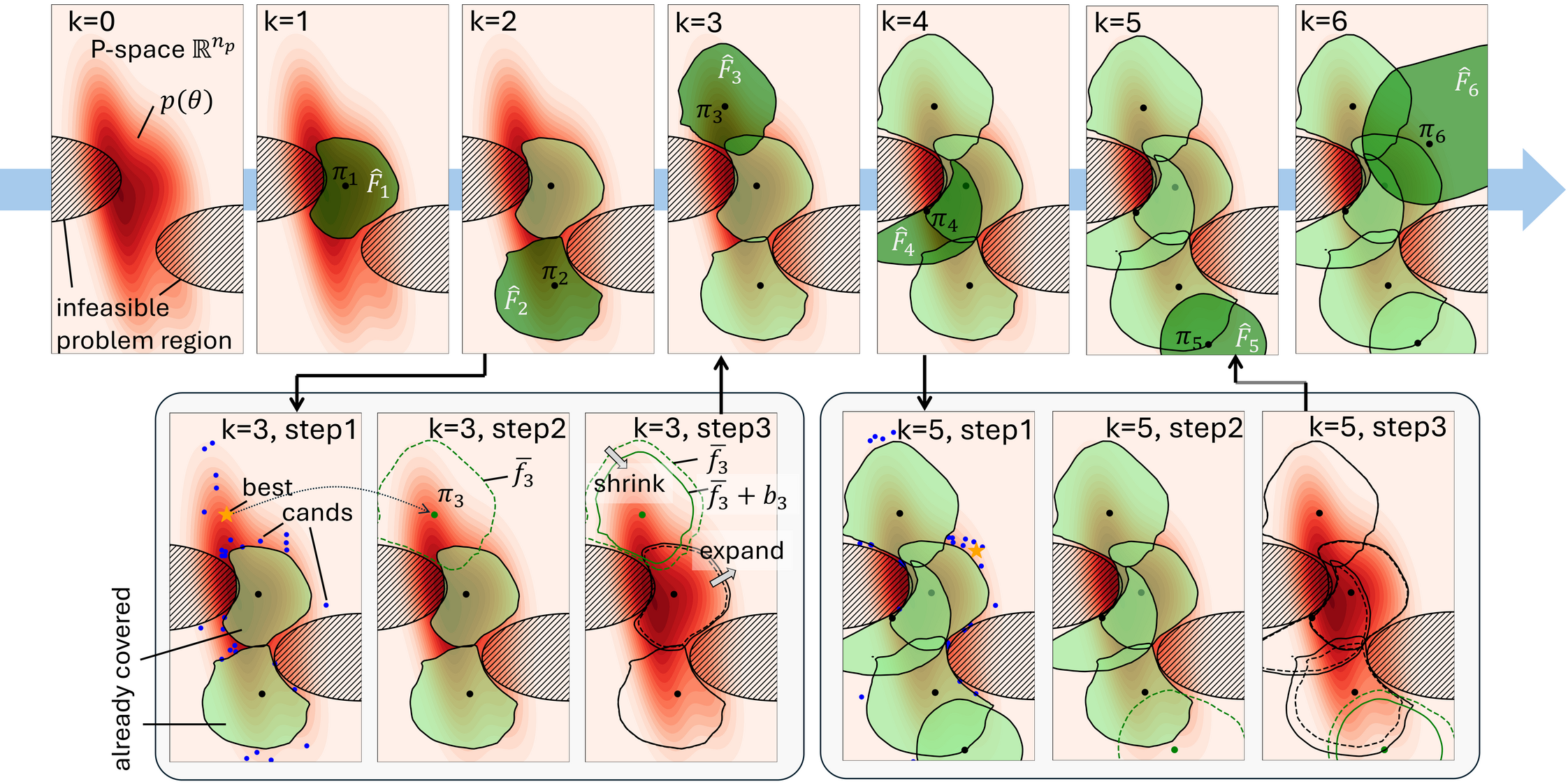}
    \vspace{-2mm}
    \caption{
The proposed algorithm actively constructs a library to maximize coverage of $p(\theta)$. For visualization purposes, we assume a one-to-one correspondence between each problem and its solution (experience), thereby allowing experiences (e.g., best, cands, $\pi_k$) to be visualized within the P-space (note that this correspondence is not generally the case).
    Iterations $k=3$ and $k=5$ are highlighted, and steps 1 to 3 for each iteration are illustrated. Step 1 selects the next experience $\pi_k$, which is expected to yield the maximum coverage gain among the candidates. Step 2 trains the base term $\bar{f}_k$ via regression. Step 3 determines the bias terms $b_{1:k}$ through optimization to meet the FP rate constraint.
    During the optimization at step 3 of the $k$-th iteration, regions defined by the base term $\bar{f}_k$ (green dashed line) and by previously determined regions (black dashed lines) are either shrunk or expanded to the regions depicted respectively by green and black solid lines.
    }
    \vspace{-3mm}
    \label{fig:sequence}
\end{figure*}

\section{Iterative Algorithm for library construction}
This section presents an iterative algorithm, CoverLib, to approximately solve the library construction problem (\ref{eq:main_problem}).

\subsection{Method overview of CoverLib}
Problem (\ref{eq:main_problem}) is challenging due to its combinatorial black-box nature and functional optimization aspect. The combinatorial black-box nature arises from the exponential growth of the search space as the number of experiences increases and the need for Monte Carlo (MC) integration to estimate $r(\cdot)$ and $\alpha(\cdot)$. This makes it intractable to exploit the problem's structure for optimization. To address this combinatorial black-box challenge, we decompose the problem into a sequence of subproblems, thereby forming a greedy iterative algorithm. 

The functional optimization aspect presents another challenge, which stems from $\hat{f}_{1:k}$ being the functions that form optimization variables.
To address this functional optimization challenge, we decompose $\hat{f}_i$ into $\hat{f}_i = \bar{f}_i + b_i$, where $\bar{f}_i$ is the highly parameterized \textit{base} term and $b_i \in \mathbb{R}$ is the \textit{bias} term. Here, $+$ can denote the sum of a function and a scalar, treating the scalar as a constant function. We obtain the base term through regression on a dataset generated by $(\theta, c) \sim p(c|\theta, \pi_i)p(\theta)$, treating only $b_{1:k}$ as the optimization variables in (\ref{eq:bias_determination}).

The adoption of the greedy iteration above is motivated by a well-known fact: the greedy algorithm achieves an efficient $(1-1/e)$-approximation of the optimal solution when the objective function is submodular \cite{nemhauser1978analysis, krause2014submodular}. Assuming perfect classifiers \footnote{This also requires the adaptation algorithm to be deterministic.} obtained instantly without any data-generation/training and a finite search space for experiences, the library construction problem (\ref{eq:main_problem}) is equivalent to the weighted MAX-COVER problem, a well-known example of submodular function maximization. The structural similarity between (\ref{eq:main_problem}) and the weighted MAX-COVER problem suggests that the greedy approach is effective in practice.

Each iteration consists of three steps: (step 1) heuristically determining $\pi_k$ that is expected to have a large \textit{coverage gain} $r(\hat{f}_{1:k}) - r(\hat{f}_{1:k-1})$, which is estimated using the information of $\hat{f}_{1:k-1}(=\{\bar{f}_i + b_i\}_{i=1}^{k-1})$; (step 2) training the base term $\bar{f}_k$ using the dataset generated based on $\pi_k$; (step 3) determining the bias terms $b_{1:k}$ by the following optimization:
\begin{equation}\label{eq:bias_determination}
\max_{b_{1:k}} r(\left\{ \bar{f}_i + b_i \right\}_{i=1}^{k}),\; \mathrm{s.t.}\; \alpha(\pi_{1:k}, \left\{ \bar{f}_i + b_i\right\}_{i=1}^{k} ) \leq \delta
\end{equation}
where $\pi_{1:k}$ and $\bar{f}_{1:k}$ are fixed in the optimization. Note that the bias terms can be both positive and negative, which respectively make the classification conservative or optimistic. While the first and second steps do not consider the FP rate constraint, the third step does and compensates for its absence in the first two steps.
The variable dependencies in each step are roughly as follows:
\begin{align*}
\bar{f}_{1:k-1}, b_{1:k-1} &\rightarrow \pi_k, \;&\text{(step 1, line 4 of Alg.\;\ref{algo:algo_whole})}\\
\pi_k &\rightarrow \bar{f}_k, \;&\text{(step 2, lines 5-6 10-11 of Alg.\;\ref{algo:algo_whole})}\\
\pi_{1:k}, \bar{f}_{1:k} &\rightarrow b_{1:k}. \;&\text{(step 3, line 7 of Alg.\;\ref{algo:algo_whole})}
\end{align*}

The overall procedure is summarized in Algorithm \ref{algo:algo_whole} and an illustration of the algorithm is shown in Fig.\;\ref{fig:sequence}. The while loop with the breaking condition $\hat{r}_{\mathrm{opt}} > \hat{r}_{\mathrm{pre}}$ is necessary because each iteration does not guarantee an increase in coverage. The following subsections will provide a more detailed explanation of each of the three steps. 

\SetKwRepeat{Do}{do}{while}%
\newcommand{\gray}[1]{\textcolor{red}{#1}}
\newcommand{\nmc}{N_{\mathrm{mc}}}
\newcommand{\mcset}{\{\theta_{\mathrm{test}}^{(n)} \}_{n=1}^{N_{\mathrm{mc}}}}
\newcommand{\fhatset}[1]{\{\hat{f}_i\}_{i=1}^{#1}}
\newcommand{\fbarset}[1]{\{\bar{f}_i\}_{i=1}^{#1}}
\newcommand{\bias}{b}
\begin{algorithm}[t]
    \label{algo:algo_whole}
    \caption{CoverLib algorithm}
    \nonl \textbf{procedure:} $\mathrm{buildLibrary}(c_{\mathrm{max}}, \delta, K)$ \\
    \nonl \textbf{tuning param: $\gamma_{\mathrm{init}}$} \\
    \small
    $\hat{r}_{\mathrm{pre}} \leftarrow 0, \gamma \leftarrow \gamma_{\mathrm{init}}$ \\
    \SetInd{0.1em}{1em}
    \For{$k \in 1..K$}{
        \SetInd{0.1em}{1em}
        \Do{$\hat{r}_{\mathrm{opt}} \leq \hat{r}_{\mathrm{pre}}$}{
            $ \pi_k, \hat{\Delta}(\pi_k) \leftarrow \mathrm{determineExperience}(\hat{f}_{1:k-1}, \maxit)$ \\
            $ n_{\mathrm{data}} \leftarrow \gamma / \hat{\Delta}(\pi_k)$ \\
            $ \bar{f}_k \leftarrow \mathrm{trainModel}(\pi_k, n_{\mathrm{data}})$ \\
            $ b_{1:k}, \hat{r}_{\mathrm{opt}} \leftarrow \mathrm{determineBiases}(\pi_{1:k}, \bar {f}_{1:k}, \delta, \maxit) $ \\
        }
        $\hat{f}_i(\theta) \xleftarrow{def} \bar{f}_i(\theta) + b_i, \forall i \in \{1..k\} $ \\
        $\rho \leftarrow (\hat{r}_{\mathrm{opt}} - \hat{r}_{\mathrm{pre}}) / \hat{\Delta}(\pi_{k-1})$ \\
        update $\gamma$ following the rule (\ref{eq:gamma_update}) \\
        $\hat{r}_{\mathrm{pre}} \leftarrow \hat{r}_{\mathrm{opt}}$\\

    }
    \textbf{return} $\pi_{1:K}, \hat{f}_{1:K}$
\end{algorithm}

\subsection{Next experience $\pi_k$ determination (step 1)} \label{sec:coverlib_step1}
This procedure corresponds to $\mathrm{determineExperience}$ in Alg.\;\ref{algo:algo_whole} and is detailed in Alg.\;\ref{algo:determine_traj}. The goal is to identify the next experience $\pi_k$ that will maximize coverage gain. To accomplish this, we follow the process: first, we generate $n_\mathrm{cand}$ candidate experiences $\Pi_{\mathrm{cand}}$. Then, for each candidate $\pi \in \Pi_{\mathrm{cand}}$, we calculate the approximate coverage gain $\Delta(\pi)$ in (\ref{eq:gain_est}) using MC integration and select the candidate yielding the highest gain.

The approximate coverage gain $\Delta(\pi)$ represents the probability that random variables $(\theta, c)$ meet two criteria: $\theta$ is not currently covered by any previous experiences ($\theta \notin \cup_{i=1}^{k-1} \hat{F}_{i}$) and can be solved using experience $\pi$ within $c \leq \maxit$. This is formally expressed as:
\begin{equation}\label{eq:gain_est}
\Delta(\pi) := \prob{(\theta, c)}{\theta \notin \cup_{i=1}^{k-1} \hat{F}_{i} \land c \leq \maxit}
\end{equation}
where $(\theta, c) \sim p(c|\theta, \pi)\cdot p(\theta)$.
This value $\Delta(\pi)$ serves as an approximation of the true coverage gain $r(\hat{f}_{1:k}) - r(\hat{f}_{1:k-1})$. This is because direct computation of the true gain is not possible since $\hat{f}_{1:k}$ will be determined in subsequent steps 2 and 3. Our choice of maximizing $\Delta(\pi)$ is reasonable because it becomes equivalent to maximizing the true coverage gain when $\hat{f}_{1:k}$ achieves perfection. 

It is important to note that $p(c|\theta, \pi)$ is not provided analytically but rather is procedurally defined via an actual adaptation process. More precisely, the random variable pair $(\theta, c)$ is sampled via the following procedure: sample $\theta \sim p(\theta)$, then adapt $\pi$ to $\theta$ and measure the computational cost $c$.

The determination of $\pi_{k}$ is performed by $\argmax_{\pi} \Delta(\pi)$. This optimization is carried out by random search (RS) with $n_{\mathrm{cand}}$ candidate experiences. 
In the RS, the choice of the search distribution of $\pi$ is crucial. The adopted search distribution is induced by the following procedures: sample $\theta \sim p(\theta)$ until $\theta \notin \cup_{i=1}^{k-1} \hat{F}_{i}$ (e.g. by sampling by rejection), then solve for such $\theta$ using a from-scratch planner and obtain $\pi$ as the solution. The motivating heuristic behind this choice of the search distribution is the following: \textit{Consider a problem $\theta$ and one of its solutions $\pi$. If a new problem $\theta'$ is similar to $\theta$, then adaptation of $\pi$ to $\theta'$ takes fewer computational costs.} This heuristic motivates confining the search space to the uncovered region $(\cup_{i=1}^{k-1} \hat{F}_{i})^{c}$, because such a solution to a problem in the uncovered region is expected to be easily adapted to other problems in the same region.

In the first whole loop of Alg.\;\ref{algo:determine_traj}, the candidate set $\Pi_{\mathrm{cand}}$ of the RS is sampled from the search distribution.
The second while loop samples $n_{\mathrm{mc}}$ uncovered problems $\Theta_{\mathrm{uncovered}}$ for MC estimation of $\Delta$.
The final for loop computes the expected gain and selects the best experience $\pi_{\mathrm{best}}$. Note that the $\mathrm{fromScratchPlan}(\theta)$ procedure returns the solution path as an experience if successful; otherwise, it returns nil.
It is recommended to use a probabilistic complete planner with sufficient timeout for $\mathrm{fromScratchPlan}$.

\begin{algorithm}[h]
    \label{algo:determine_traj}
    \caption{Determine next experience to add}
    \small
    \nonl \textbf{procedure:} $\mathrm{determineExperience}(\hat{f}_{1:k-1}, \maxit)$\\
    \nonl \textbf{tuning param:} $(n_{\mathrm{cand}}, n_{\mathrm{mc}}) = (100, 500)$, these values are fixed throughout this article. \\
    \tcp{Determined candidate experiences $\Pi_{\mathrm{cand}}$}
    $\Pi_{\mathrm{cand}} \leftarrow \{\}$ \\
    \While{ $\# \Pi_{\mathrm{cand}} < n_{\mathrm{cand}}$}{
        $\theta \sim p(\theta)$ \\
        \If{$\theta \notin \cup_{i=1}^{k-1} \hat{F}_{i}$}{
            $\pi \sim \mathrm{fromScratchPlan}(\theta)$ \\
            \If{$\pi \neq nil$}{
                $\Pi \leftarrow \Pi_{\mathrm{cand}} \cup \{\pi\}$ \\

            }
        }
    }
    \tcp{Sample uncovered problems $\Theta_{\mathrm{uncovered}}$}
    $\Theta_{\mathrm{uncovered}} \leftarrow \{\}, n_{\mathrm{total}} \leftarrow 0$ \\
    \While{ $\# \Theta_{\mathrm{uncovered}} < n_{\mathrm{mc}}$}{
        $n_{\mathrm{total}} \leftarrow n_{\mathrm{total}} + 1$ \\
        $\theta \sim p(\theta)$ \\
        \If{$\theta \notin \cup_{i=1}^{k-1} \hat{F}_i$}{
            $\Theta_{\mathrm{uncovered}} \leftarrow \Theta_{\mathrm{uncovered}} \cup \{\theta\}$ \\
        }
    }

    \tcp{Select the best experience $\pi_{\mathrm{best}}$}
    $\hat{\Delta}_{\mathrm{best}} \leftarrow 0, \pi_{\mathrm{best}} \leftarrow nil$ \\
    \For{$\pi \in \Pi_{\mathrm{cand}}$}{
        adapt $\pi$ to all $\Theta_{\mathrm{uncovered}}$ then count the problems that are successfully adapted within $\maxit$ as $successCount$ \\
        $\hat{\Delta} \leftarrow successCount / n_{\mathrm{total}}$ \\
        \If{$\hat{\Delta} > \hat{\Delta}_{\mathrm{best}}$}{
            $\pi_{\mathrm{best}} \leftarrow \pi, \hat{\Delta}_{\mathrm{best}} \leftarrow \hat{\Delta}$ \\
        }
    }
    \textbf{return} $\pi_{\mathrm{best}}, \hat{\Delta}_{\mathrm{best}}$
\end{algorithm}

\subsection{Regression of base term $\bar{f}_k$ (step 2)} \label{sec:step2}
This step determines the base part $\bar{f}_k$ of $\hat{f}_k$ by regression. The mean squared error between the predicted and actual computational cost serves as the loss function for regression. Although any model can be adopted for $\bar{f}_k$, as of 2024, neural networks are likely the most suitable choice due to their expressive power and tool availability. The dataset for regression is generated by sampling $n_{\mathrm{data}}$ times from the distribution $(c, \theta) \sim p(c|\theta, \pi_k) \cdot p(\theta)$.

Sampling $n_{\mathrm{data}}$ from $p(c|\theta, \pi_k)$, as mentioned in Section \ref{sec:coverlib_step1}, involves solving the motion planning problem $n_{\mathrm{data}}$ times. As $n_{\mathrm{data}}$ increases, so does the computational cost for regression. Therefore, minimizing $n_{\mathrm{data}}$ is essential for efficiency. However, model accuracy must be balanced against computational cost. An inaccurate model from insufficient $n_{\mathrm{data}}$ data will necessitate a larger bias term $b_k$ in step 3 to meet the FP rate constraint, yielding a conservative coverage gain compared to the actual gain.

One key observation for adaptively determining $n_{\mathrm{data}}$ is the following: \textit{Assume the adaptation algorithm is deterministic for simplicity \footnote{$R_k$ cannot be defined for stochastic adaptation algorithms.}. Let $R_k$ be a region where adaptation of $\pi_k$ is successful but not those of $\pi_{1:k-1}$. Then the number of samples in the dataset that fall into $R_k$ is proportional to $\Delta(\pi_k)$ in (\ref{eq:gain_est}).} This observation suggests that as the iteration proceeds, the number of samples falling into the region $R_k$ will gradually decrease if $n_{\mathrm{data}}$ remains constant throughout the iterations. To maintain a sufficient number of samples in such a region to preserve accuracy, $n_{\mathrm{data}}$ should be inversely proportional to the coverage gain $\Delta(\pi_k)$ estimated in step 1; that is $n_{\mathrm{data}} = \gamma/\hat{\Delta}(\pi_k)$ (line 5 of Alg.\;\ref{algo:algo_whole}), where $\gamma$ is an inverse proportionality constant. While this heuristic essentially substitutes the challenge of determining $n_{\mathrm{data}}$ with that of determining $\gamma$, the latter is easier to tune as it reflects the aforementioned observation.

Still, tuning $\gamma$ presents a similar trade-off as tuning $n_{\mathrm{data}}$. To address this, we adopt the following strategy: we set the initial value $\gamma_1$ at the beginning and then adaptively change it throughout the iteration according to the \textit{achievement rate} $\rho_k := (\hat{r}_{k-1} - \hat{r}_{k-2}) / \hat{\Delta}(\pi_{k-1})$, which represents the ratio of the actual coverage gain to the estimated coverage gain. A small (near 0) $\rho_k$ indicates that $n_{\mathrm{data}}$ should have been larger in the previous iteration; thus, we increase $\gamma_k$ and vice versa. We use the following update rule for $\gamma_k$ throughout the paper:
\begin{equation} \label{eq:gamma_update}
    \gamma_k = \begin{cases}
        \gamma_{k-1} \times 1.1 & \text{if $\rho_k < 0.3$} \\
        \gamma_{k-1} \times 0.9 & \text{if $\rho_k > 0.7$} \\
        \gamma_{k-1} & \text{otherwise}.
    \end{cases}
\end{equation}
This $\gamma$ update procedure corresponds to lines 10-11 of Alg.\;\ref{algo:algo_whole}.

\subsection{Bias terms $b_{1:k}$ determination procedure (step 3)}
This step determines $b_{1:k}$ by solving (\ref{eq:bias_determination}). The optimization aims to balance the trade-off between coverage gain and FP rate, as larger bias terms result in higher coverage gain but also a higher FP rate and vice versa. $r(\cdot)$ and $\alpha(\cdot)$ are evaluated via MC integrations, making it natural to employ a black-box optimization (BBO) method. As the iteration progresses, the optimization by BBO becomes more challenging due to the increased dimensionality $k$ of the optimization variables and the fact that even small differences in the objective function values become more important. Instead, a simple RS approach is often more effective than BBO for larger $k$. Thus, fixing $b_{1:k-1}$ and optimizing $b_k$ (bias for the latest trained classifier) using a simple RS method is effective. We used both approaches in the iterations and selected the better one.

We utilize the Covariance Matrix Adaptation Evolution Strategy (CMAES) \cite{hansen2003reducing}, a standard BBO method \footnote{We used the implementation of \cite{nomura2024cmaes} with the default setting other than setting $\sigma = \maxit/2$. The initial values of $b_{1:k}$ are set to all zero. The maximum iteration number is set to $1000$.}. As CMAES, like most BBO methods, cannot directly handle hard inequality constraints, we reformulate the objective function by incorporating the inequality term as a penalty:
\begin{align}
J(b_{1:k}) & = - r(\left\{ \bar{f}_i + \bias_i \right\}_{i=1}^{k}) \\
+ 10^{4} & ||\mathrm{max}(0, \delta' - \alpha(\pi_{1:k}, \{ \bar{f}i + \bias_i\}_{i=1}^{k}))||^{2} \nonumber
\end{align}
where $\delta'$ is set to $\delta - 10^{-3}$ to encourage the minimizer to satisfy the inequality constraint, albeit with increased conservativeness. However, this modification does not yet guarantee that the minimizer of $J(b_{1:k})$ will satisfy $\alpha(\cdot) \leq \delta$. To address this issue, we run CMAES 2000 times with different initial seeds, obtaining multiple minimizers of $J(b_{1:k})$ on different local optima. We then select the minimizer that satisfies $\alpha(\cdot) \leq \delta$ and achieves the highest coverage rate $r(\cdot)$.

The MC estimation of $r(\cdot)$ and $\alpha(\cdot)$ in CMAES and RS is computed by $N_{\mathrm{cover}} / \nmc$ and $N_{\mathrm{fp}} / N_{\mathrm{cover}}$, respectively, where
\begin{align}\label{eq:mc_estimation}
& N_{\mathrm{cover}} = \sum_{j=1}^{\nmc} \mathbbm{1} \{ \min_{i\in\{1..k\}}(\overline{c}^{(i, j)} + b_i) \leq \maxit \} \\
& N_{\mathrm{fp}} = \sum_{j=1}^{\nmc} \mathbbm{1} \{ i' = \argmin_{i\in\{1..k\}}(\overline{c}^{(i, j)} + b_{i}), \\
& \qquad\qquad\qquad \bar{c}^{(i', j)} + b_{i'} \leq \maxit \land c^{(i', j)} \leq \maxit \} \nonumber.
\end{align}
Here $c^{(i, j)}$ is a sampled value from $p(c|\theta^{(j)}, \pi_i)$ and $\bar{c}^{(i, j)}$ is the evaluation of $\bar{f}_i(\theta^{(j)})$. Although sampling from $p(c|\theta^{(j)}, \pi_i)$ is computationally expensive as it performs actual adaptation, it is possible to share $\{c^{(i, j)}\}_{i=1, j=1}^{k, \nmc}$ and $\{\bar{c}^{(i, j)}\}_{i=1, j=1}^{k, \nmc}$ across the CMAES and RS iterations, because only $b_{1:k}$ is optimized and the others are fixed.

We further reduce the computational cost by sharing the problem set $\Theta_{\mathrm{mc}} = \{\theta^{(j)} \}_{j=1}^{\nmc}$ across all $k\in[K]$ iterations. We can cache the values of $c^{(i, j)}$ and $\bar{c}^{(i, j)}$ for all $i \in [k]$ and $j \in [\nmc]$, and reuse them in step 3 in the subsequent iterations (i.e. $k+1\ldots$-th iterations in Alg.\;\ref{algo:algo_whole}). This caching technique reduces the total number of samples from $p(c|\theta, \pi)$ from $K(K-1)\nmc/2$ to $K\nmc$. Throughout this article, a large value of $\nmc=10^{4}$ is used to ensure the accuracy of the MC estimation, but this does not become a computational bottleneck thanks to the above caching techniques.

\subsection{Reduce computational time through large parallelization} \label{sec:massive_parallel}
The proposed algorithms may experience bottlenecks in five procedures: 1) sampling from $p(\theta)$, 2) sampling from $p(c|\theta, \pi)$ which involves the adaptation operation, 3) $\mathrm{fromScratchPlan}$ procedure, 4) $\mathrm{runCMAES}$ procedure, and 5) training the base term $\bar{f}_k$ through regression. Fortunately, the first four procedures can be easily parallelized. In our implementation, we evenly distribute these tasks across two servers (see Sec.\;\ref{sec:computer_resource}) where each server has a worker pool. Parallelizing these procedures significantly reduces the computational time. Although the last $\bar{f}$ training procedure could also be parallelized using distributed training techniques, we did not implement this in the current study.

%% file: section/setting.tex
\section{Shared settings and implementation details in numerical experiments} \label{sec:setting}

\subsection{Baseline: passive random sampling nearest neighbor search library} \label{sec:nearst_neighbor_library}
For comparison of the high-level parts of the library-based methods, we adopt a baseline approach called NNLib, which utilizes passive random sampling for library construction and nearest neighbor search for querying. As reviewed in Section \ref{sec:nearest_neighbor}, most library-based planning algorithms adopt an NNLib-style approach or its variants. The NNLib is constructed by repeating the following process: sampling $\theta \sim p(\theta)$ then solving it from scratch using a global planner with a sufficient time budget. If the problem is solved, the experienced solution $\pi$ is stored in the library with the key $\theta$. Here, for a fair comparison, the global planner and its timeout are set to be the same as in CoverLib (i.e. $\mathrm{fromScratchPlan}$).
During the online planning phase, given the P-parameter $\theta_{\mathrm{q}}$, the solution of the nearest key to $\theta_{\mathrm{q}}$ is retrieved from the library.
Our NNLibs's query implementation employs a ball-tree \cite{omohundro1989five} to enable efficient query search. Hereinafter, when we refer to the \textit{library size} of NNLibs, it represents the number of solved problems used to construct the library rather than the number of stored experiences in the library.

For the baseline, we constructed NNLib with two distinct features: NNLib(full), which uses the full dimension of $\theta$ for the query, and NNLib(selected), which uses only the initial and goal condition-related components of $\theta$. Examples of such components include the initial base pose and reaching target coordinates. This feature selection is motivated by previous research \cite{berenson2012robot, pairet2021path}, which shows that using only the initial and goal conditions as the query remains effective even if the environment changes. As NNLib is expected to be susceptible to the curse of dimensionality, such a feature selection can potentially improve its performance.

\subsection{On Islam et al.'s approach as a potential baseline} \label{sec:islams_work_baseline}
The work \cite{islam2019provable, islam2021provably} by Islam et al. reviewed in Section \ref{sec:islams_work} is another potential baseline method. Although applicable to all settings in the numerical experiments with minor modifications, we omitted it from the comparison for the following reasons. Their approach requires discretization. Even in Domain 2 (see Section \ref{sec:humanoid_bench}), with $n_{p}=10$ being the lowest among the Domains in Section \ref{sec:experiment}, the computational cost would be significant even with a coarse discretization (e.g., 10 bins per dimension). Suppose a single adaptation takes 0.1 seconds and we consider using the two \armservers (in total 156 cores) stated in Section \ref{sec:computer_resource} for fully parallel processing. Checking the adaptability of the selected experience path for all $10^{10}$ possible problems would take $10^{10} \cdot 0.1 / 156 / 3600 / 24 = 74$ days at the first iteration. Repeating a similar \footnote{From the next iteration, the adaptability check for already covered problems can be skipped. Thus, the computational cost will be reduced over iterations; however, an order of magnitude similar to the first iteration is expected.} process many times to populate the library would require an immense amount of computational time. The computational challenge worsens exponentially for Domains 1, 3 and 4, which have dozens or hundreds of dimensions. This is why we did not consider their approach in the comparison.

\subsection{Adaptation algorithms} \label{sec:adaptation_algorithms}
For both NNLib and CoverLib, we are free to choose the low-level part, that is, the adaptation algorithm. In this paper, we consider the following two adaptation algorithms from different categories: one from the a) NLP-based category and another from the b) SBMP-based category. Algorithm a) is a general-purpose algorithm, while algorithm b) is more suitable for unconstrained path planning problems; thus, we applied them accordingly. We describe the specific implementation details of the algorithms (SQPPlan and ERTConnect) corresponding to both categories in the following.

\textbf{a) Sequential Quadratic Programming based planning (SQPPlan):}
Most motion planning problems can be formulated as constrained NLPs. Adaptation of a specific experience $\pi$ can be done straightforwardly in this framework by solving the optimization problem with $\pi$ as the initial guess of the solution. We adopt sequential quadratic programming (SQP) as the optimization algorithm. The computation cost $c$ for SQPPlan is measured by the outer iteration count, not the iteration count of the internal QP solver. To exploit the sparse block structure in the trajectory optimization, we use OSQP \cite{stellato2020osqp} as the internal QP solver. Note that our SQP implementation is simplified; line-search and trust-region are omitted, and the Hessian of the constraints is ignored.

\textbf{b) Experience-driven random tree connect (ERTConnect):}
Experience-driven random tree connect (ERTConnect) \cite{pairet2021path} is the state-of-the-art sampling-based adaptation algorithm. In addition to its speed and broad adaptability, ERTConnect excels in path quality without requiring path simplification, unlike the approach used in Lightning \cite{berenson2012robot}. Therefore, we used ERTConnect without any path simplification. As ERTConnect is a bidirectional algorithm, it requires determining the path's terminal configuration in advance. Determining the terminal configuration from scratch is computationally expensive, as it involves solving the constrained inverse kinematics (IK) with multiple initial guesses, which often takes a comparable amount of time to the path planning itself. To address this, our implementation takes advantage of the terminal configuration of the retrieved experience as the initial guess. If the IK failed with that initial guess, the entire planning process was considered a failure. The constrained-IK is implemented using scipy's \cite{virtanen2020scipy} implementation of SQP (referred to as scipy-SQP). The computation cost $c$ for ERTConnect is measured by the number of collision checks inside the path-planning algorithm ignoring the ones in the IK solver.

\subsection{Modeling and regression of the base term $\bar{f}$} \label{sec:nn_architecture}
\begin{figure}[t]
    \centering
    \includegraphics[width=0.95\linewidth]{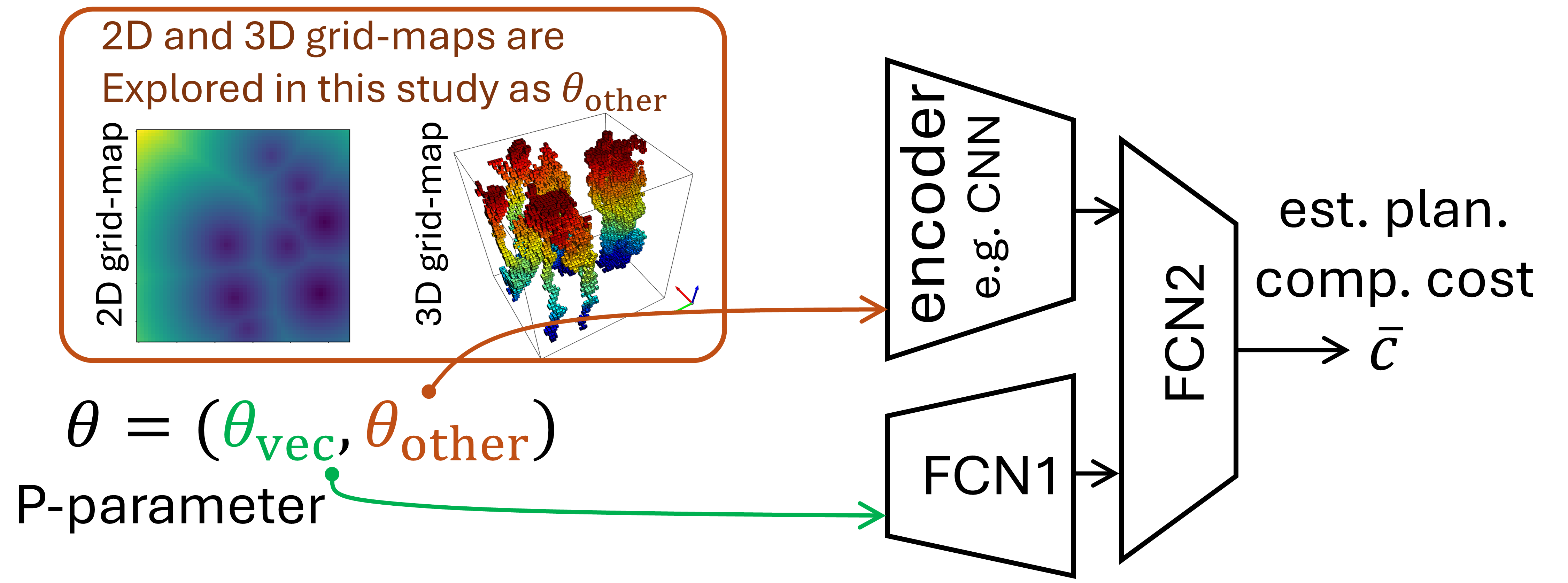}
    \vspace{-2mm}
    \caption{Illustration of vector-encoder modeling.}
    \vspace{-5mm}
    \label{fig:vector_matrix_modeling}
\end{figure}

We model the base term $\bar{f}$ using a neural network architecture. We consider two types of modeling: \textit{vector modeling} and \textit{vector-encoder modeling}. Vector modeling is used when the P-parameter is simply a real vector. This approach uses a single fully connected neural network (FCN). The FCN takes a vector-formed P-parameter $\theta$ as input and outputs the expected value $\bar{c}$ of the computational cost. Alternatively, the vector-encoder modeling is used when a part of the P-parameter is not a vector but a more complex structure, such as a grid map. In this case, we split the P-parameter $\theta$ into the vector part $\theta_{\mathrm{vec}}$ and the non-vector part $\theta_{\mathrm{other}}$ (see Fig.\;\ref{fig:vector_matrix_modeling}). In this modeling, the non-vector part is encoded into a vector via the encoder network. The vector part $\theta_{\mathrm{vec}}$ is, first, processed by the FCN1 to enlarge the dimension to match the order of its dimension to that of the encoded counterpart. Then, the outputs of both networks are concatenated and processed by the FCN2. In this paper, we studied the case where parts of the P-parameter are 2D and 3D arrays, and we used corresponding 2D and 3D CNNs. However, other modeling approaches, such as PointNet or Graph CNN, are also possible and exploring them is an interesting future direction.

In all experiments adopting the vector-encoder modeling, we first pre-trained the encoder part of the vector-encoder model through self-supervised learning before running Algorithm \ref{algo:algo_whole}. The encoder remained fixed during the training of $\bar{f}$, with only the FCN components being updated. This approach of using a fixed pre-trained encoder offers great benefit in both training and inference phases. Specifically, in the inference time in Algorithm \ref{algo:algo_whole}, the encoder is evaluated only once, rather than $K$ times for all $\bar{f}_{1:K}$. This modeling strategy also significantly reduces memory consumption, as the encoder component typically requires more memory than the FCN part.

For the pre-training details, we employed 2D CNN encoders for Domains 1 and 3, and a 3D CNN encoder for Domain 4. In each case, we implemented an hourglass-shaped CNN autoencoder trained through self-supervised learning. While reconstruction loss was used as the primary training metric, Domain 4 required special consideration. When working with binary occupancy maps, we found that standard autoencoder training was unsuccessful. This occurred because elements where $occ=0$ typically outnumber elements where $occ=1$, causing the reconstruction output to become trapped in a local minimum that produces an all-near-zero reconstruction output. To solve this challenge, we defined the loss function using the difference between outputs and the L1 unsigned distance grid map of input voxels. Due to the efficient ``two-pass algorithm" \cite{rosenfeld1968distance}, the voxel-to-L1 distance transformation does not create major computational bottleneck.

For more details on the network architectures, training, and inference settings, please refer to Appendix.

\subsection{Computational resources} \label{sec:computer_resource}
For the experiments, we utilized two \armservers (ARM Neoverse N1 80-Core, 256GB DRAM, 3.3GHz) and a single \desktop (AMD Ryzen 9 3900X 12-Core with SMT, 64GB DRAM, 3.8GHz, GeForce RTX 2080 Ti). Please note that the Neoverse N1 is a server-grade processor and is not for consumer use. For pretraining the encoder part of the vector-encoder model, we used solely \desktop. Subsequently, we used both \armservers and \desktop for the library construction phase.  As stated in \ref{sec:massive_parallel}, most of the time-consuming procedures in library construction, except for fitting $\bar{f}$, can be parallelized. Hence, we performed parallelizable computations on each \armserver using 78 parallel processes (-2 reserved for other purposes), totaling 156 processes. In contrast, during the benchmarking phase, only \desktop was used, with a single physical core (i.e., 2 logical cores) isolated from the Linux process scheduling.

%% file: section/benchmark.tex
\section{Numerical Experiment} \label{sec:experiment}
\subsection{Benchmarking Methodology} \label{sec:benchmarking_methodology}
We conduct numerical experiments to evaluate the proposed method in four different motion planning domains, focusing on the following two aspects:

\textbf{Planning Performance Comparison (PPC):} We compare the performance of CoverLib against three other category of methods: global planners (referred to as \textit{Global}), NNLib(full) and NNLib(selected). The Global is supposed to be a probabilistically complete algorithm, meaning that given enough time, it can find a solution if one exists. While CoverLib and NNLibs are not expected to achieve the same level of success rate as Global, our aim is to evaluate how closely they can approach this upper bound while requiring less planning time.

Note that the success rate does not often provide a comprehensive picture of a method's ability to handle difficult problems. In a typical domain, the majority of problems are often easy, while a smaller subset of problems, usually in specialized or niche areas of the P-space, are considerably more challenging. As a result, the overall success rate may be skewed towards the performance on easier problems.
To better gauge the capability of each method, we introduce a new metric called Global Planner Equivalent Timeout (GPE-TO). GPE-TO represents the timeout value at which Global achieves the same success rate as the method being evaluated. To calculate GPE-TO, we first preprocess the benchmark results of Global to create a mapping between timeout values and their corresponding success rates (see plots for Global in Fig.\;\ref{fig:tradeoff}). Then, for each method, we find the timeout value in this mapping that corresponds to the method's success rate.
For each domain, we benchmark the methods using 3000 (Domains 1 and 2) or 1000 (Domains 3 and 4) problems sampled from $p(\theta)$. Then, we compare three key aspects: GPE-TO, success rate, and planning time.

\textbf{Library Growth Comparison (LGC):} Alongside the PPC, we perform a Library Growth Comparison (LGC) to assess the evolution of CoverLib and NNLib's success rates as their libraries grow. While PPC provides a snapshot of the methods' performance at a specific library size, LGC offers a comprehensive picture of efficiency in library construction by plotting the success rate growth for each library-based method as the library construction time increases. Both NNLib's and CoverLib's success rate growth curves are computed by measuring the actual success rate in the same manner as in the PPC with different library sizes. 

In the plots for both PPC and LGC, the approximate upper bound of the success rate of the domain is indicated by a red horizontal line. This upper bound is estimated by running Global with a sufficiently large (180 sec) timeout for 3000  problems randomly sampled from $p(\theta)$. It's important to note that this upper bound does not reach 100\% even with an infinite timeout in general, as some problems in the domain can be infeasible.

\subsection{Domain 1: Kinodynamic planning for double integrator} \label{sec:double_integrator}

\textbf{Domain definition ($n_{\mathrm{p}} = 32$):}
Two dimensional double integrator is a simple dynamical model described by a linear system $\ddot{x} = u$, where $x \in \mathbb{R}^2$ is the position and $u \in \mathbb{R}^2$ is the acceleration control input.
We consider a $[0, 1] \times [0, 1]$ square-shaped world with 10 randomly placed $\in \mathbb{R}^{10\times 2}$ circular obstacles of random radii $\in \mathbb{R}^{10}$.
The start $v_{\mathrm{start}}$ and goal $v_{\mathrm{goal}}$ velocities are both fixed at $(0, 0)$. The velocity and acceleration are bounded by $||\dot{x}||_{\infty} \leq 0.3$ and $||u||_{\infty} \leq 0.1$. The goal position $x_{\mathrm{goal}} \in \mathbb{R}^{2}$ is randomly sampled but the start position is fixed $x_{\mathrm{goal}} = (0.1, 0.1)$. Therefore, in total the DOF of $p(\theta)$ is $n_{\mathrm{p}} = 32$. The P-space is defined as the Cartesian product of $\mathbb{R}^{2}$ and $\mathbb{R}^{56 \times 56}$. The former represents the goal position and the latter represents the $56 \times 56$ grid map of the signed distance field (SDF) of the obstacles in the environment. The kinodynamic (kinematic + dynamic) motion planning problem for the double integrator is defined as finding a sequence of control inputs $u_1, u_2, \ldots, u_{N-1}$ that satisfies the following constraints:
\begin{align}
    & x_1 = x_{\mathrm{start}}, \dot{x}_1 = \dot{x}_{\mathrm{start}}, x_N = x_{\mathrm{goal}}, \dot{x}_N = \dot{x}_{\mathrm{goal}} & \label{eq:bubbly_boundary_condition} \\
    & \dot{x}_{i+1} = \dot{x}_i + u_i \Delta t, \quad \forall i \in [N-1] & \label{eq:bubbly_dynamics_eq1} \\
    & x_{i+1} = x_i + \dot{x}_i \Delta t + u_i \Delta t^2/2, \quad \forall i \in [N-1] & \label{eq:bubbly_dynamics_eq2} \\
    & \text{sdf}(x_i) \geq 0, \quad \forall i \in [N] & \label{eq:bubbly_collision_ineq} \\
    & x_i \in [0, 1] \times [0, 1], ||\dot{x}_i||_{\infty} \leq 0.3, \quad \forall i \in [N] & \label{eq:bubbly_velocity_ineq} \\
    & ||u_i||_{\infty} \leq 0.1, \quad \forall i \in [N-1] & \label{eq:bubbly_acceleration_ineq}
\end{align}
where $\Delta t$ is the time step and we set $\Delta t = 0.3$. $N$ is the number of discretization points. While $N$ is fixed for trajectory optimization, it varies for sampling-based planners. The $\mathrm{sdf}$ is the signed distance function (SDF).

\textbf{Adaptation algorithm:}
We adopt SQPPlan as the adaptation algorithm. The trajectory optimization problem is constructed by direct transcription \cite{tedrake2009underactuated} with $N=200$ discretization points with time step $\Delta t$. The cost function is the sum of squared control input: $\sum_{i=1}^{N-1} ||u_i||^2$ and the constraints are given by (\ref{eq:bubbly_boundary_condition})-(\ref{eq:bubbly_acceleration_ineq}).

\begin{figure*}[t]
    \centering
    \includegraphics[width=0.98\linewidth]{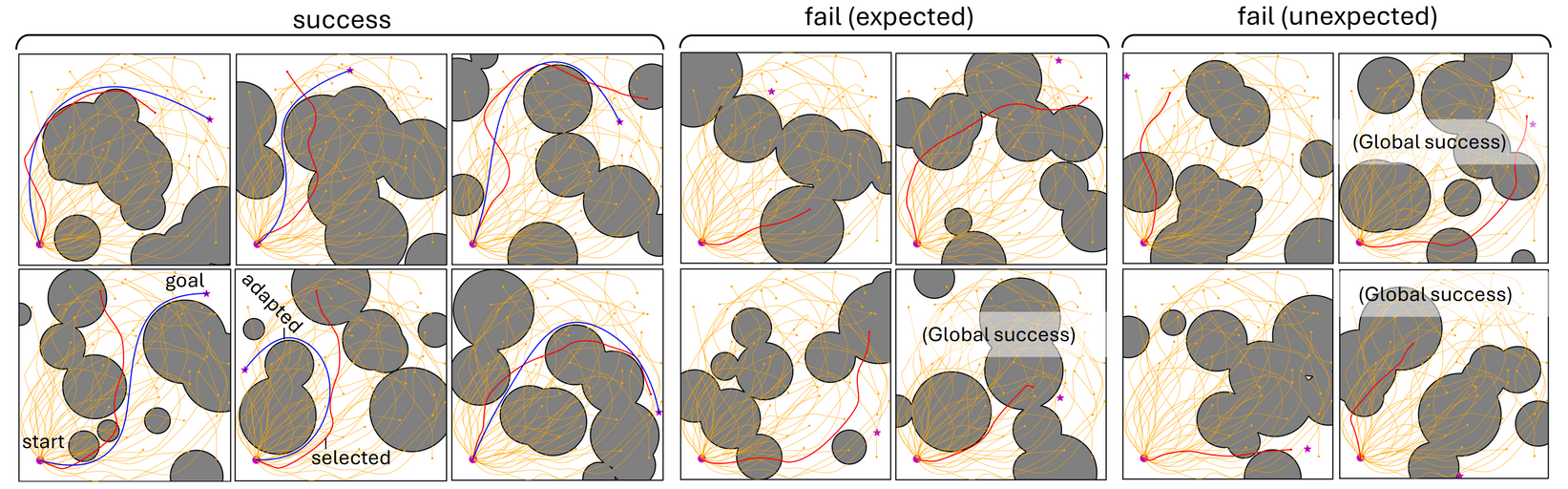}
    \vspace{-2mm}
    \caption{This figure visualizes the behavior of the CoverLib-aided planning with $(c_{\mathrm{max}}, \delta)  = (5, 0.05)$ in Domain 1. The \textcolor{Dandelion}{Yellow lines} are the experiences in the library, while the \textcolor{Red}{red line} is the selected experience used in adaptation. The \textcolor{Blue}{blue line} shows the result of adaptation. The \textcolor{DarkOrchid}{purple circle and star} denote the start and goal positions respectively. The blue line is not shown for failure cases as the result is not obtained. The label ``fail(expected)" indicates cases where $\mathrm{min}_{i \in [K]} \hat{f}_i(\theta) > \maxit$, meaning failure is somewhat expected before adaptation. The label ``fail(unexpected)" signifies that adaptation failed even though $\mathrm{min}_{i \in [K]} \hat{f}_i(\theta) \leq \maxit$. The overlaid "Global success" label for failed cases suggests that the problem is actually feasible, as evidenced by the success of Global.}
    \vspace{-3mm}
    \label{fig:bubbly_visualization}
\end{figure*}
\begin{figure}[t]
    \centering
    \includegraphics[width=1.0\linewidth]{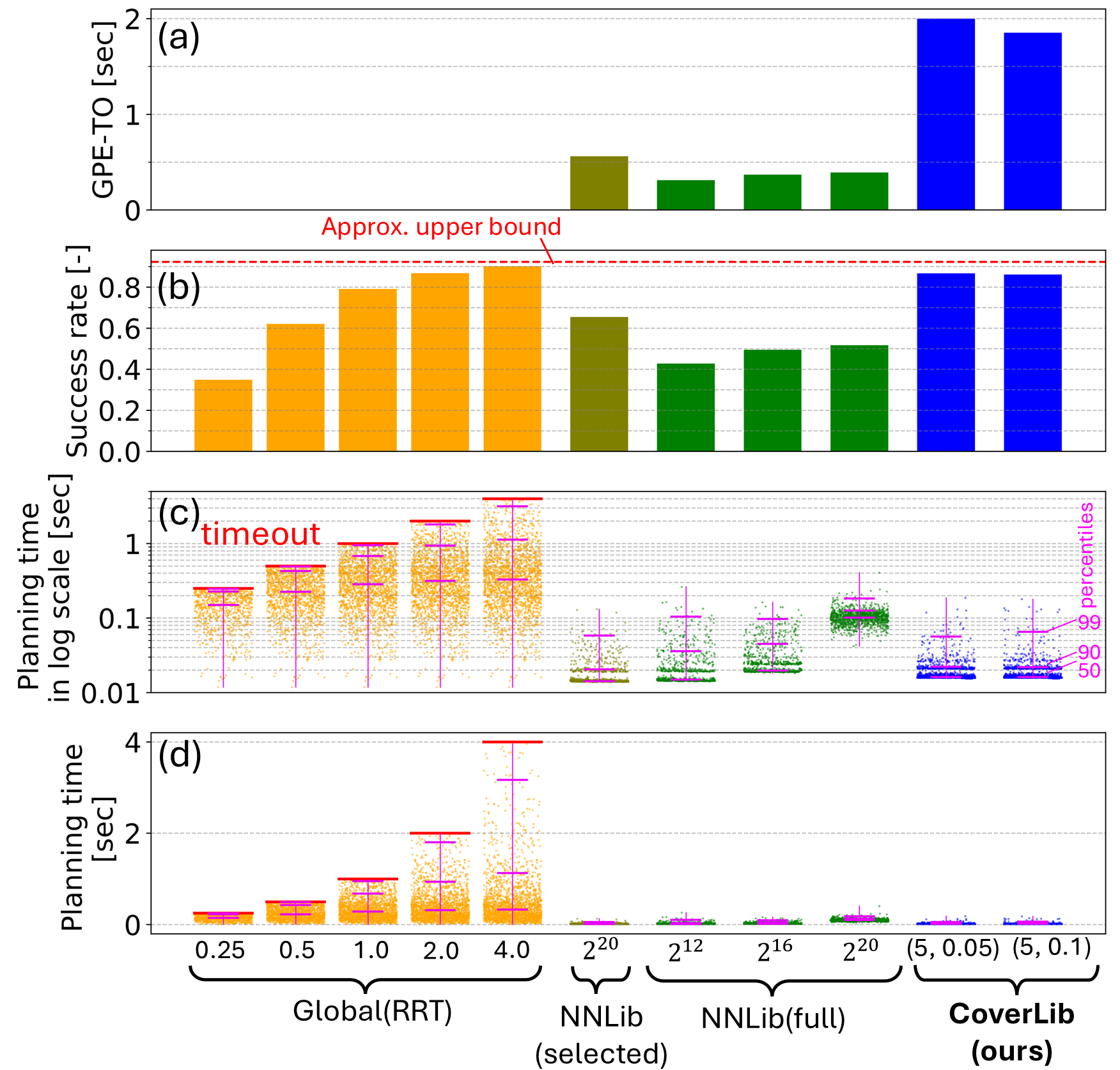}
    \vspace{-5mm}
    \caption{
        Benchmarking in terms of (a) GPE-TO, (b) Success rate, (c-d) Planning time for Domain 1. The values associated with the Global plot (e.g. 0.25, 0.5, ...) represent timeout values, which are depicted by short red bars in (c-d). The value associated with NNLib (e.g. $2^{20}$) indicates the library size. The values associated with CoverLib, e.g. (5, 0.05), are $(\maxit, \delta)$ pairs. In (c) and (d), the x-coordinate of each point in the scatter plot represents the problem index, while the y-coordinate represents the corresponding planning time. Data points for problems with failure are not plotted. The vertical pink lines indicate the minimum and maximum planning time of the scatter plot. The three horizontal pink lines along each scatter plot indicate the 50, 90, and 99 percentiles of the planning time. These percentile computations only consider successful cases. The approx. upper bound in (b) is computed as described in Section \ref{sec:benchmarking_methodology}.  
    }
    \label{fig:result_double_integrator1}
    \vspace{-5mm}
\end{figure}
\begin{figure}[t]
    \centering
    \includegraphics[width=1.0\linewidth]{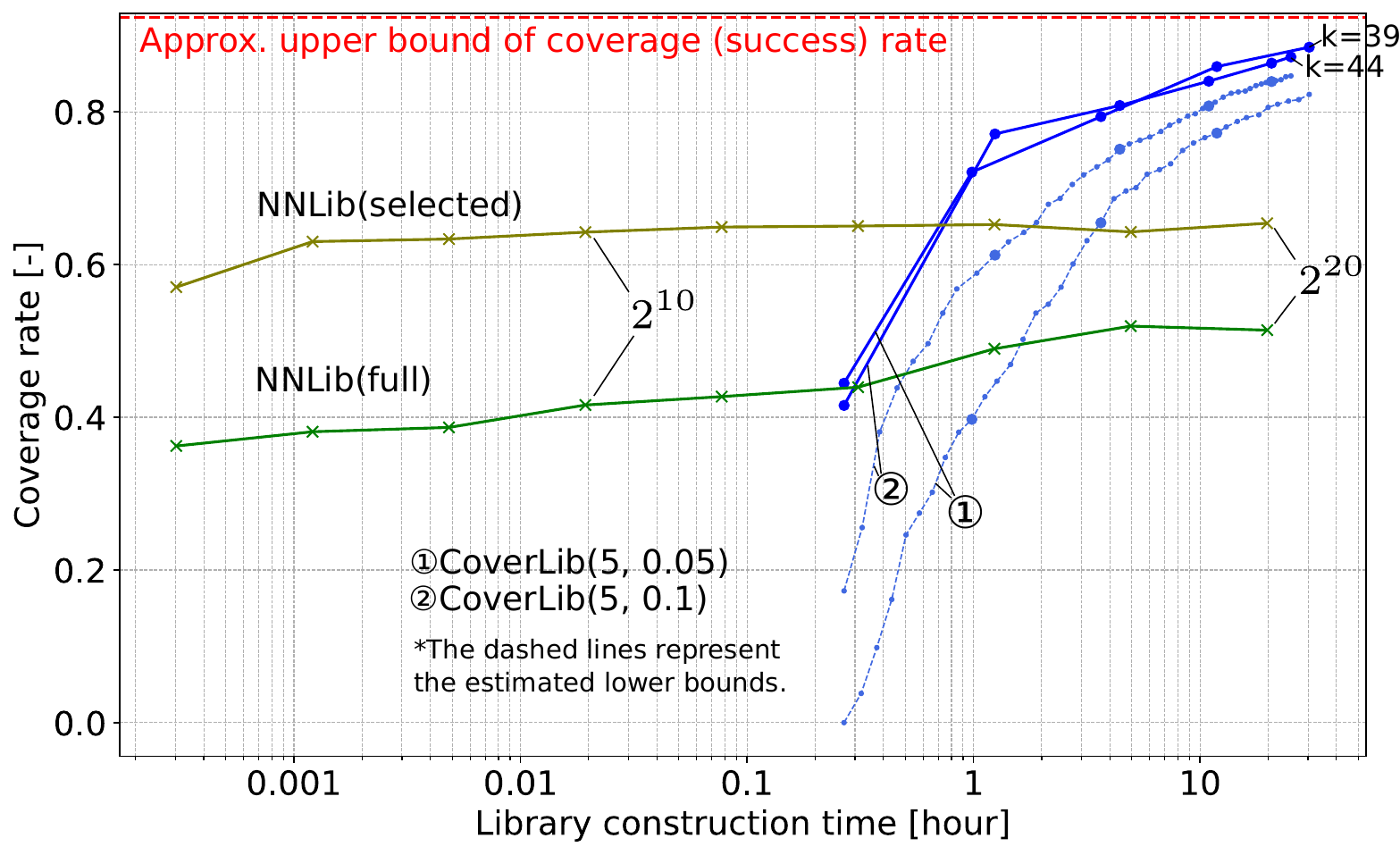}
    \vspace{-3mm}
    \caption{Comparison of library growth for Domain 1. For NNLib, the library size varies as $2^i$ for $i \in$ (4, 6, 8, 10, 12, 14, 16, 17, 18, 19, 20), where $2^{20} = 1,048,576$. The corresponding library construction time for each specific library size is calculated from the logging results. To ensure a fair comparison, the data generation for NNLib is performed using the two \armservers mentioned in Section \ref{sec:computer_resource}. For CoverLib, measurements are plotted at iterations 1, 10, 20, and every subsequent increment of 10 until the last iteration. Additionally, as auxiliary information, we show the estimated lower bound of the coverage rate $(1 - \delta) \cdot \hat{r}_{\mathrm{opt}}$ for each iteration $k$ with a dashed light blue line, where every 10th point is emphasized with a larger marker. The approximate upper bound is computed as described in Section \ref{sec:benchmarking_methodology}. \textbf{Note that the pre-training time for the CNN encoder is included in the library construction time, which results in the initial library construction time being shifted to the right by the pre-training time.}
    }
    \vspace{-5mm}
    \label{fig:result_double_integrator2}
\end{figure}

\textbf{From-scratch planner:}
We adopt the fast marching tree (FMT$^*$) \cite{janson2015fast} for $\mathrm{fromScratchPlan}$, where an analytical time-optimal connection is used to connect the nodes \cite{webb2013kinodynamic}. The FMT$^*$ is run with a sufficiently high number of samples, $N=3000$, to find a feasible path. Single planning by this setting typically takes several seconds.

\textbf{Cost predictor:}
We adopt the vector-encoder modeling for $\bar{f}$. The encoder part here is 2D CNN, which takes the $56 \times 56$ grid map of the SDF as input. The CNN encoder is pre-trained as described in Section \ref{sec:nn_architecture}. The pre-training time including dataset generation was about 13 minutes on \desktop.

 \textbf{Result:} The library is constructed with $(c_{\mathrm{max}}, \delta) = (5, 0.1)$ with about 1 day budget and $(5, 0.05)$ within about 1.5 days budget. Tuning parameter $\gamma$ is set to 5000. The behavior of planning with the learned CoverLib is visualized in Fig.\;\ref{fig:bubbly_visualization}. The CoverLib with SQP adaptation is benchmarked against the baselines: Global, NNLib(full), and NNLib(selected).
We adopt RRT as Global. However because straightforward extension of RRT to kinodynamic planning \cite{lavalle2001randomized} results in a quite slow planning time, we adopt an analytical time-optimal connection \cite{webb2013kinodynamic} between the nodes as did for the scratch planner. The results of PPC and LGC are shown in Fig.\;\ref{fig:result_double_integrator1} and Fig.\;\ref{fig:result_double_integrator2} respectively.

\subsection{Domain 2: kinematic planning for 31-DOF legged humanoid robot}  \label{sec:humanoid_bench}

\begin{figure}[t]
    \centering
    \includegraphics[width=0.95\linewidth]{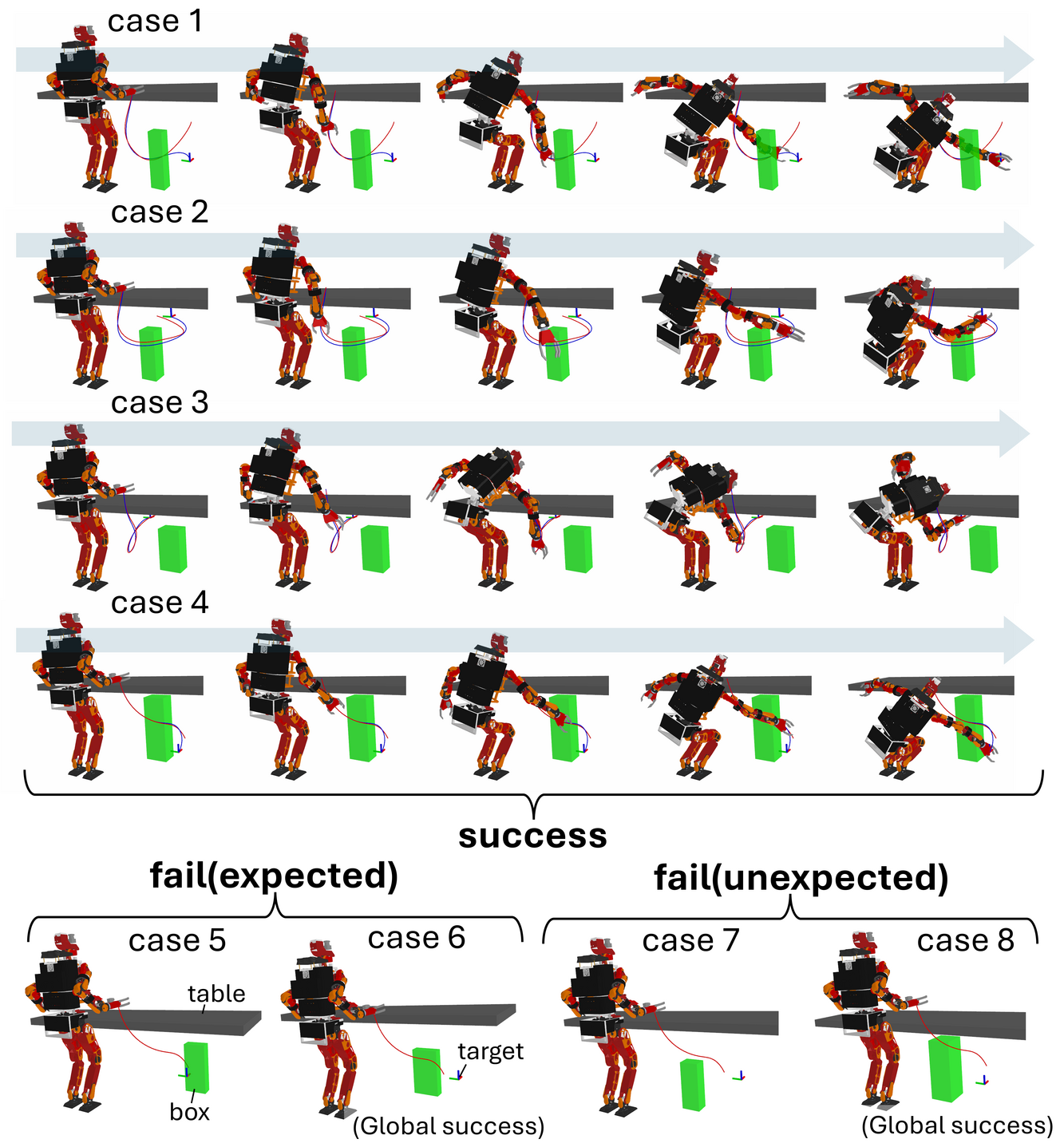}
    \vspace{-3mm}
    \caption{
        This figure visualizes the behavior of the CoverLib-aided planning with $(\maxit, \delta) = (5, 0.1)$ in Domain 2. The \textcolor{red}{red line} indicates the selected experiences from the library. The \textcolor{Blue}{blue line} is the result of adaptation. These lines are obtained by mapping the configurations to the end effector positions. See the caption of Fig.\;\ref{fig:bubbly_visualization} for the meaning of fail(expected), fail(unexpected), and (Global success) labels.
    }
    \vspace{-4mm}
    \label{fig:humanoid_visualization}
\end{figure} 
\begin{figure}[t]
    \centering
    \includegraphics[width=1.0\linewidth]{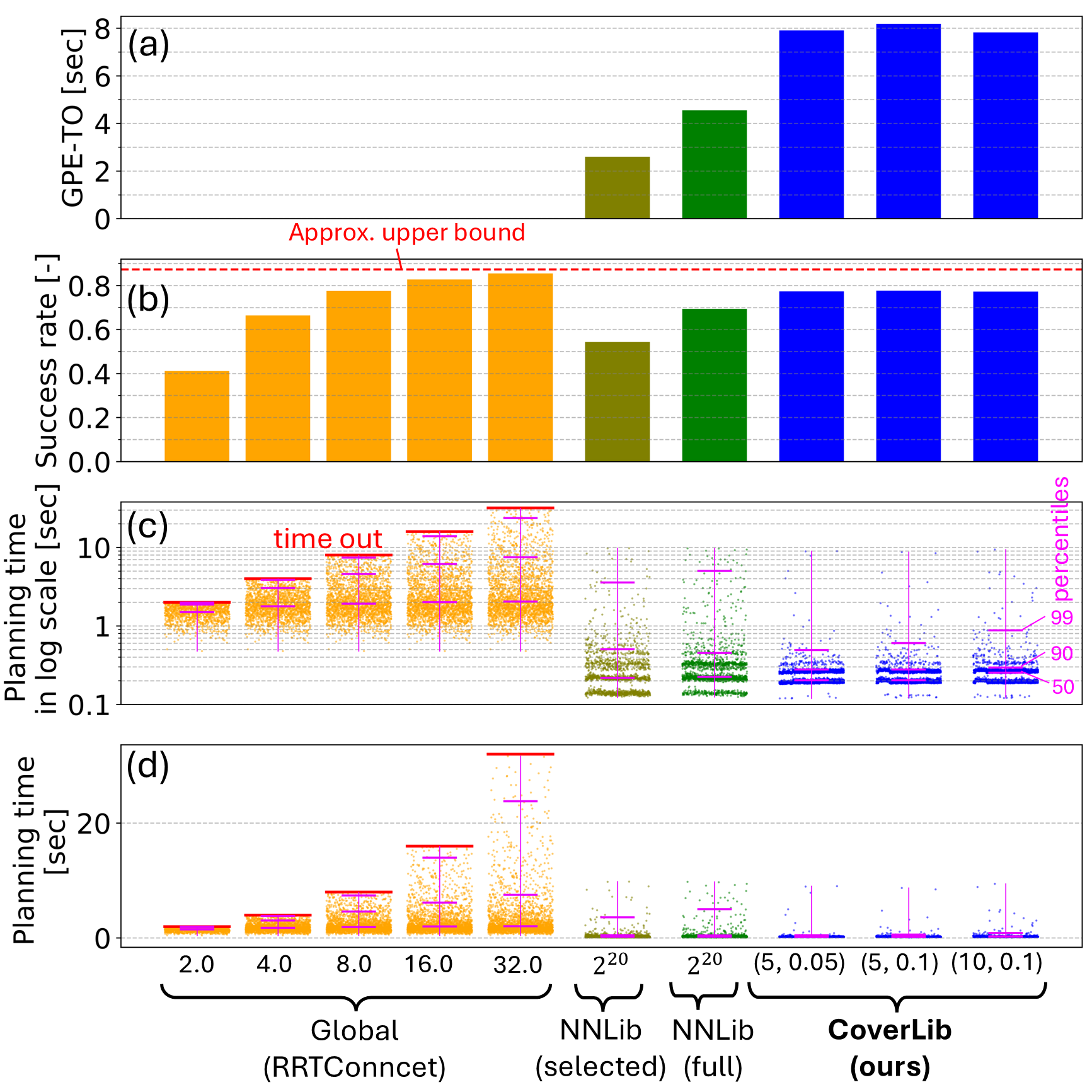}
    \vspace{-3mm}
    \caption{Planning performance comparison for Domain 2. For caption details, refer to Fig.\;\ref{fig:result_double_integrator1}.
    }
    \vspace{-3mm}
    \label{fig:result_humanoid1}
\end{figure}

\begin{figure}[t]
    \centering
    \includegraphics[width=1.0\linewidth]{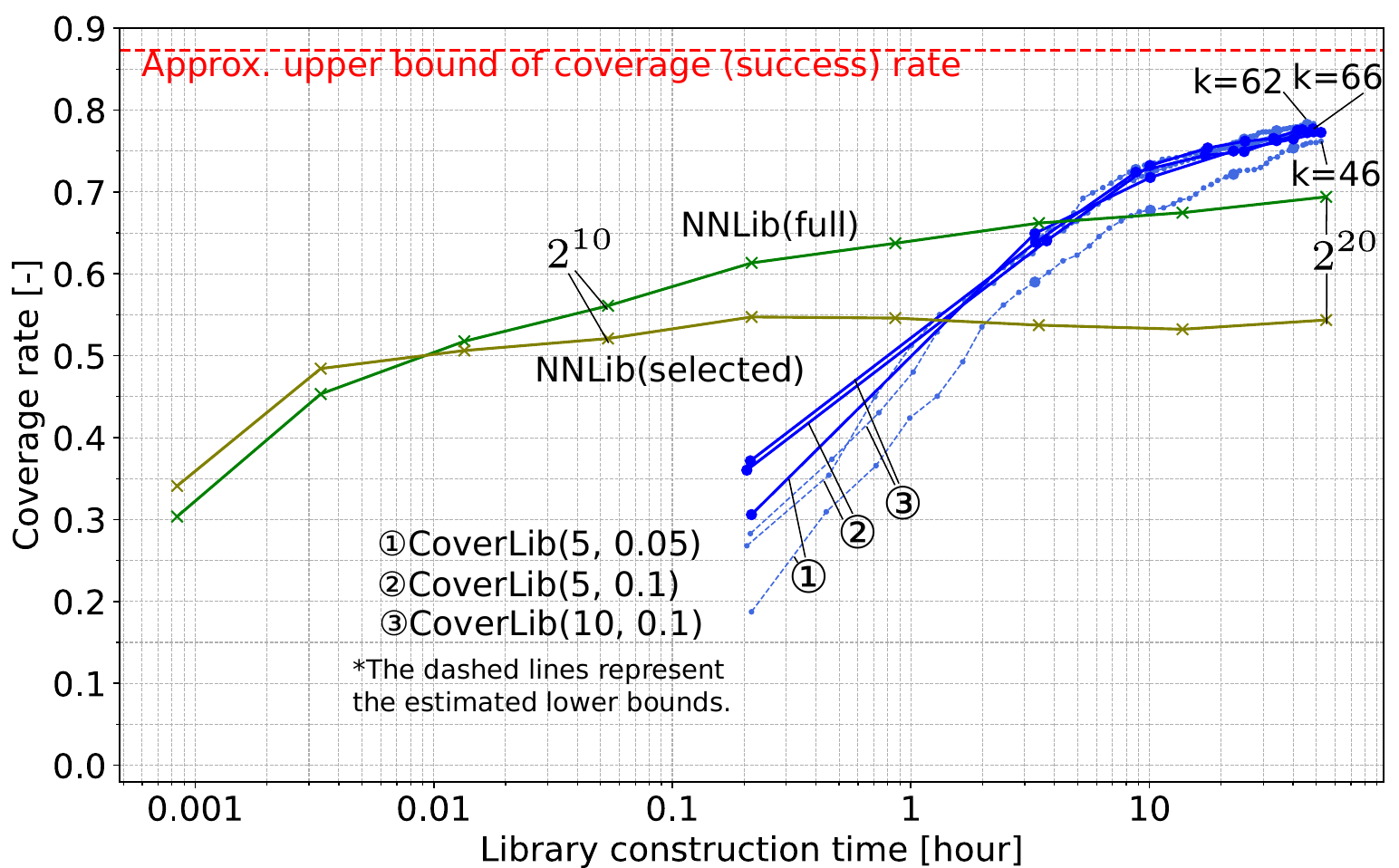}
    \vspace{-3mm}
    \caption{Library growth comparison for Domain 2. For caption details, refer to Fig.\;\ref{fig:result_double_integrator2}.
    }
    \vspace{-6mm}
    \label{fig:result_humanoid2}
\end{figure}

\textbf{Domain definition ($n_{\mathrm{p}} = 10$):}
The humanoid robot we consider is a 31-DOF (arms + legs + torso) robot JAXON \cite{kojima2015development}.
We consider the fully kinematic motion planning problem for the humanoid robot.
The robot collision shape is approximated as a set of many spheres covering the robot body. Then, the collision check is efficiently done by SDF evaluation at each sphere position, similar to \cite{zucker2013chomp, kappler2018real}.
We consider a task where the robot reaches its hand to the target position while avoiding the following two obstacles.
A fixed-size floating table-like obstacle is placed in the workspace, and its x and z positions $\in \mathbb{R}^2$ are randomly sampled.
Also, a box-shaped obstacle is placed under the ``table". The obstacle shape $\in \mathbb{R}^3$ and its planar position $\in \mathbb{R}^2$ are randomly sampled.
The target reach position $x_{\mathrm{reach}} \in \mathbb{R}^3$ is also randomly sampled in the workspace. Therefore, the DOF of $p(\theta)$ is $n_{\mathrm{p}} = 10$. The P-space here is simply $\mathbb{R}^{10}$. The motion planning problem is defined as finding a sequence of robot configurations $q_{1:L}$ that satisfies the following constraints:
\begin{align}
    & q_{1} = q_{\mathrm{start}} & \label{eq:start_eq} \\
    & endEffectorPosKin(q_{L}) = x_{\mathrm{reach}} & \label{eq:goal_eq} \\
    & f_{\mathrm{sdf}}(spherePosKin_j(q_i)) > 0, \forall i \in [L], j \in [N_{\mathrm{sphere}}] & \label{eq:collision_ineq} \\
    & q_{i} \in JointLimit, \forall i \in [L] & \label{eq:joint_limit_ineq} \\
    & comPos2dKin(q_i) \in supportPolygon, \forall i \in [L] & \label{eq:support_ineq} \\
    & leftLegKin(q_i) = (x_l, Q_1), \forall i \in [L] & \label{eq:left_leg_eq} \\
    & RightLegKin(q_i) = (x_r, Q_1), \forall i \in [L] & \label{eq:right_leg_eq} \\
    & ||q_{i+1, j} - q_{i, j}|| \leq \lambda_j, \forall i \in [L-1], \forall j \in [N_{\mathrm{joint}}] & \label{eq:pairwise_dist_ineq}
\end{align}
Note that (\ref{eq:support_ineq}) enforces the ground-projected center of mass of the robot to be inside the support polygon to prevent the robot from falling down. This constraint is treated as an inequality constraint. (\ref{eq:left_leg_eq}) and (\ref{eq:right_leg_eq}) enforce the left and right foot to be on the ground at specific coordinates $(x_l, Q_1)$ and $(x_r, Q_1)$ respectively. (\ref{eq:pairwise_dist_ineq}) enforces the pairwise distance of the each joint to be less than $\lambda_j$, which essentially defines the collision check resolution.

\textbf{Adaptation algorithm:}
We adopt the SQPPlan for the adaptation algorithm. The objective of the NLP problem is to minimize the sum of squared joint accelerations $\sum_{i=2}^{L-1} (q_{i+1} - 2 q_i + q_{i-1})^{2}$ while satisfying constraints (\ref{eq:start_eq})-(\ref{eq:pairwise_dist_ineq}). The length of the trajectory $L$ is set to 40. 

\textbf{From-scratch planner:}
We adopt a RRTConnect planner for $\mathrm{fromScratchPlan}$ with the following modifications.
The equality constraints (\ref{eq:left_leg_eq}) and (\ref{eq:right_leg_eq}) make the C-space no longer a simple Euclidean space, but a constrained manifold. Planning on a constrained manifold is known to be challenging and is an active research field \cite{kingston2018sampling}.
Basically, we use an existing projection-based approach \cite{kingston2018sampling}. However, the existing approach solely targets satisfying the equality constraints in the projection.
This often results in infeasibility in terms of the inequality constraints, and ends up repeating the projection process in vain as discussed in our recent article \cite{hiraoka2024sampling}. This problem typically occurs in humanoid planning problems, where severe inequality constraints are imposed, such as the support polygon constraint.
To satisfy both equality and inequality constraints, we adopt a projection operator that satisfies both of these constraints using SQP.
The goal configuration for RRTConnect is determined by a scipy-SQP-based IK that takes into account the constraints (\ref{eq:goal_eq})-(\ref{eq:right_leg_eq}).
We applied SQPPlan adaptation (the adaptation algorithm works as a path smoother given a non-smooth path) to the RRTConnect's solution to smooth it out. The time budget for IK + RRTConnect + smoothing is 180 seconds in total.

\textbf{Cost predictor:}
As the P-space is a multi-dimensional real space, we simply adopt the vector modeling for $\bar{f}$.

\textbf{Result:}
The CoverLibs are constructed with multiple settings $(\maxit, \delta) = (5, 0.05), (5, 0.1), (10, 0.1)$ with approximately 2 days budgets. The tuning parameter $\gamma$ is set to 10000.

The performance is compared with the baselines Global, NNLib(full), and NNLib(selected), where Global is RRTConnect, the same as the $\mathrm{fromScratchPlan}$, but without smoothing. The behavior of planning with the learned CoverLib is visualized in Fig.\;\ref{fig:humanoid_visualization}. The results of the PPC and LGC are shown in Fig.\;\ref{fig:result_humanoid1} and Fig.\;\ref{fig:result_humanoid2} respectively.

\subsection{Domain 3: Kinematic planning for 10-DOF mobile manipulator}  \label{sec:pr2_bench}
\begin{figure}[t]
    \centering
\includegraphics[width=0.98\linewidth]{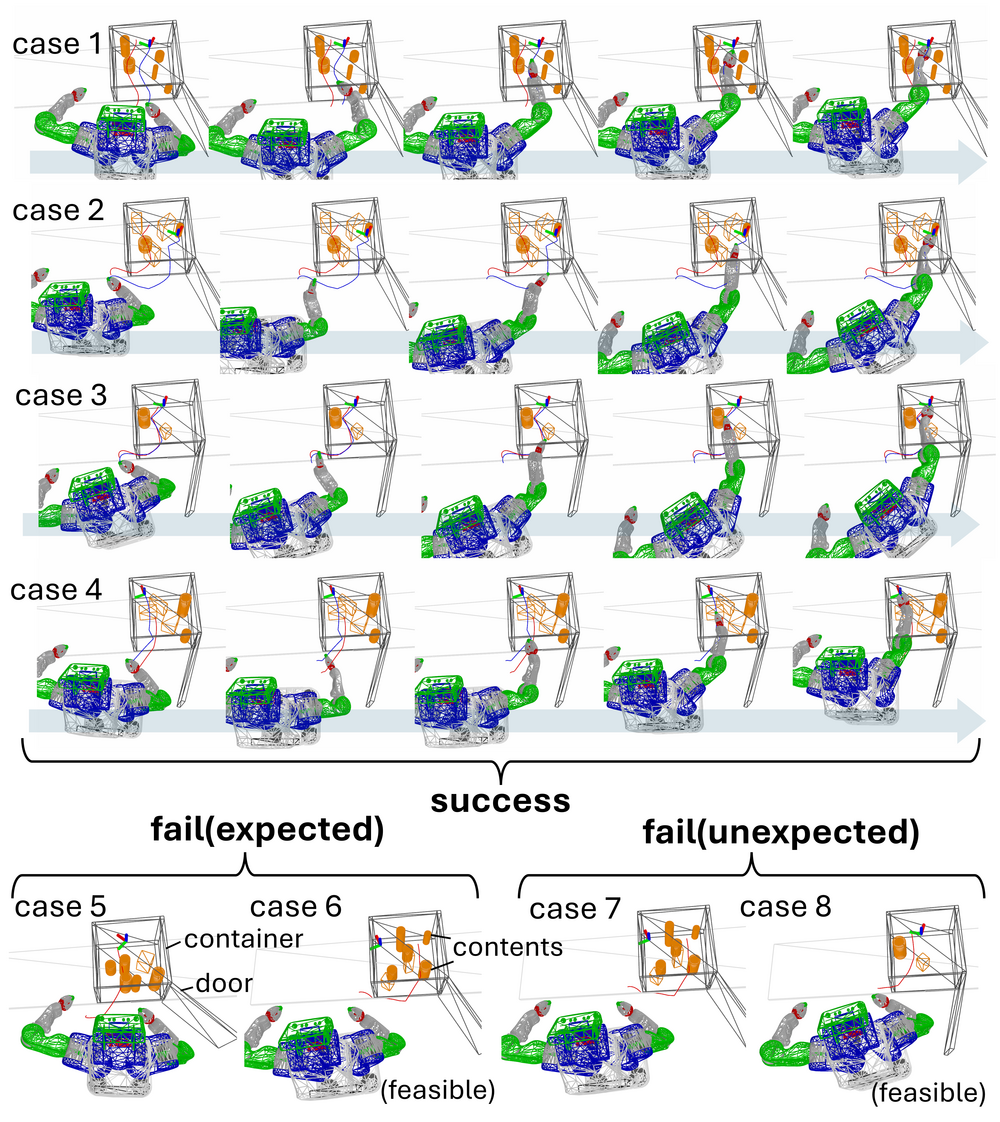}
    \vspace{-5mm}
    \caption{This figure visualizes the behavior of the CoverLib-aided planning with $(\maxit, \delta) = (2000, 0.2)$ in Domain 3. For caption details, refer to Fig.\;\ref{fig:humanoid_visualization}.}
    \label{fig:pr2_visualization}
\end{figure} 
\begin{figure}[t]
    \centering
    \includegraphics[width=1.0\linewidth]{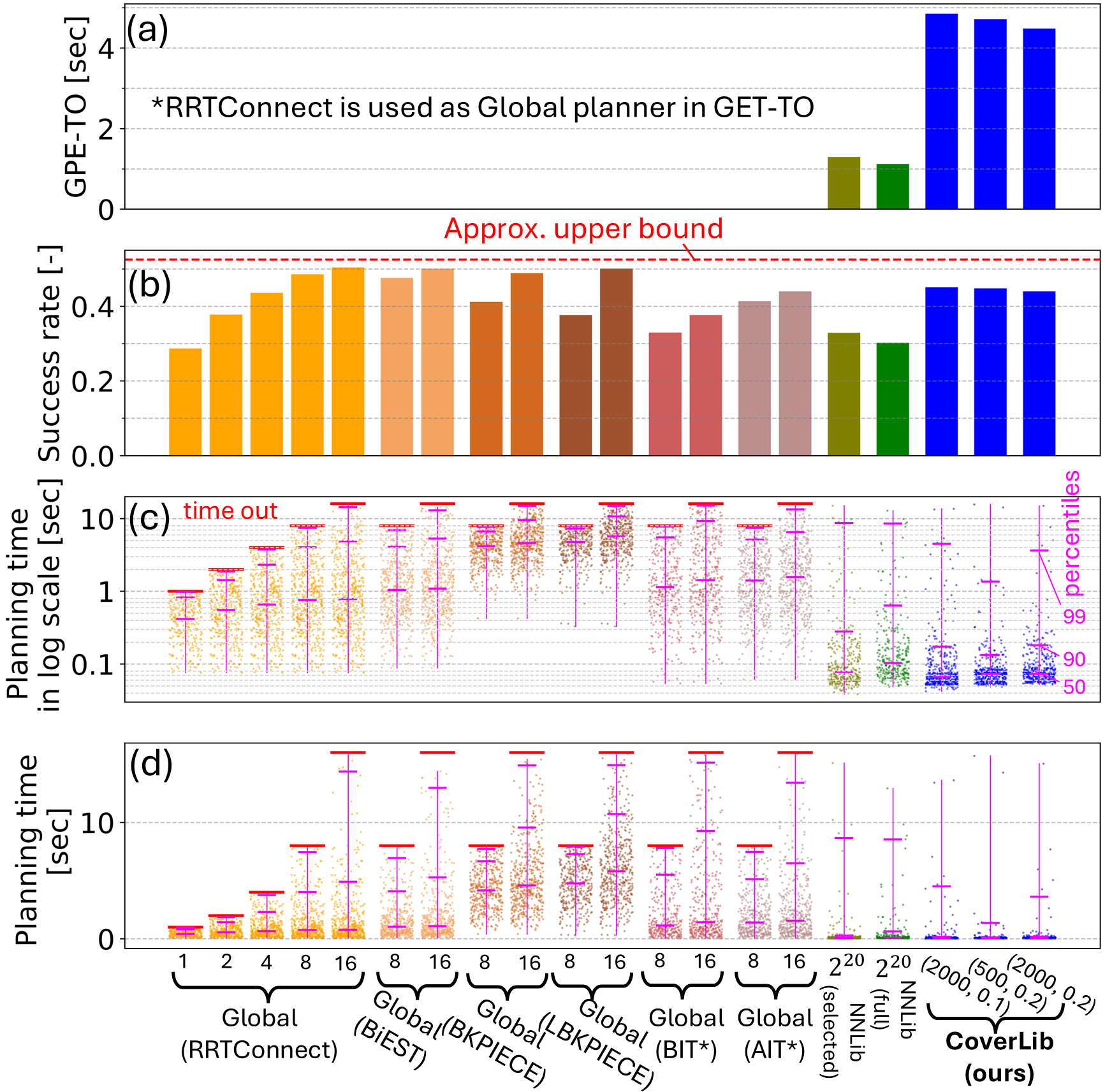}
    \caption{Planning performance comparison for Domain 3. For caption details, refer to Fig.\;\ref{fig:result_double_integrator1}.}
    \label{fig:result_pr2}
\end{figure}
\begin{figure}[t]
    \centering
    \includegraphics[width=1.0\linewidth]{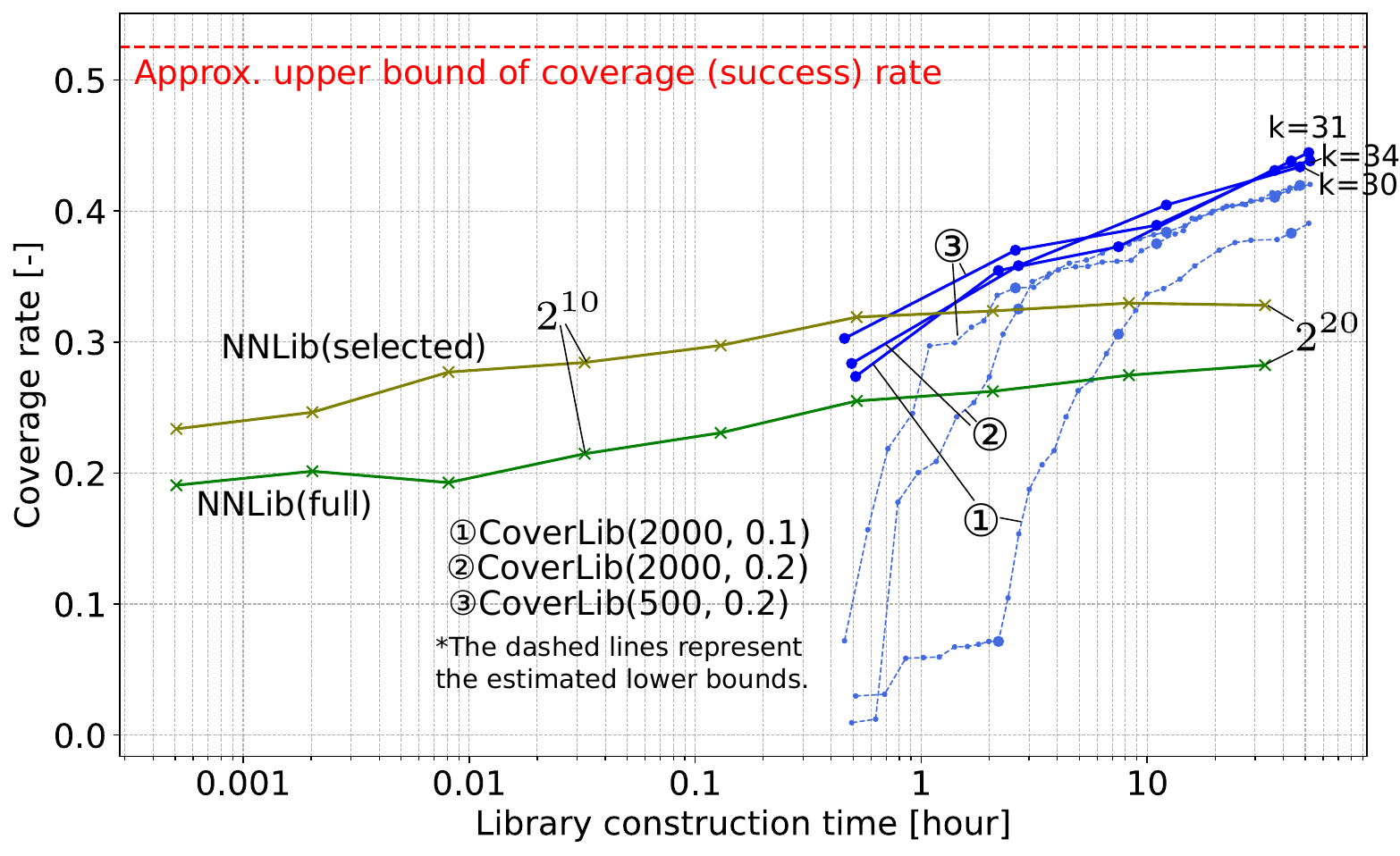}
    \caption{Library growth comparison for Domain 3. For caption details, refer to Fig.\;\ref{fig:result_double_integrator2}.}
    \label{fig:result_pr22}
    \vspace{-5mm}
\end{figure}
\textbf{Domain definition ($n_{\mathrm{p}} = 78$):}
We consider the mobile manipulator robot PR2, using the 7-DOF right arm of the PR2 and hypothetically considering the mobile base as a 3-DOF mobile base. Collision checking is done similarly to Domain 2. The task is to reach the right hand to a target position inside a container while avoiding obstacles inside the container. The container has a fixed shape, but the door angle $\in \mathbb{R}$ is randomly changed. A maximum of 10 obstacles are placed inside the container with random size and position. The obstacle is either a box or a cylinder with random size ($\in \mathbb{R}^3$ or $\in \mathbb{R}^2$) and planar coordinate ($\in \mathbb{R}^3$ or $\in \mathbb{R}^2$), where the yaw angle is considered only for the box coordinates). The number of obstacles is randomly sampled from 1 to 10. Considering a discrete variable indicating whether each obstacle is a box or a cylinder, the environmental part of DOF of $p(\theta)$ is $10 \times (1 + \max\{3 + 3, 2 + 2\}) + 1= 71$ in total. The starting base planar coordinate $\in \mathbb{R}^3$ is randomly sampled near the container. The target coordinate $\in \mathbb{R}^4$ is randomly sampled within the container, with only the x, y, z and yaw angles changed. 
Thus, the DOF for $p(\theta)$ is $n_{\mathrm{p}} = 78$ in total.
The P-space is defined as the Cartesian product of $\mathbb{R}^{8}$ and $\mathbb{R}^{56 \times 56}$. The former is the door angle, the target position, and the obstacle parameters, while the latter represents the $56 \times 56$ heightmap of the obstacles in the container. The motion planning problem is defined similarly to the humanoid problem, without the humanoid-specific constraints (\ref{eq:support_ineq}), (\ref{eq:left_leg_eq}) and (\ref{eq:right_leg_eq}).

\textbf{Adaptation algorithm:}
We adopt ERTConnect as the adaptation algorithm, with a modified \textit{malleability bound} parameter of 0.1 instead of the default 5.0 value \cite{pairet2021path}. This focuses the sampling space closer to the experienced path, rather than spanning the robot's full kinematic range. All other parameters remain unchanged, including the $\omega$ parameter, which is analogous to the \textit{range} parameter in RRTConnect. This enables fair comparison with OMPL's Global planners with default settings.

\textbf{From-scratch planner:}
We adopt the RRTConnect implementation of OMPL \cite{sucan2012open} with default settings for $\mathrm{fromScratchPlan}$. As in Section \ref{sec:humanoid_bench}, the goal configuration for RRTConnect is determined by solving collision-aware inverse kinematics (IK) using scipy-SQP until the feasible goal configuration is found. The solution is simplified by the shortcut algorithm in OMPL. The time budget for IK + RRTConnect + simplification is 30 seconds in total.

\textbf{Cost predictor:}
We adopt the vector-encoder modeling for $\bar{f}$, where the encoder part is a 2D CNN. The input to the CNN encoder is a $56 \times 56$ heightmap of the obstacles in the container. The CNN encoder is pre-trained as described in Section \ref{sec:nn_architecture}. The pre-training time including dataset generation was about 21 minutes on \desktop.

\textbf{Result:}
The CoverLibs are constructed with multiple settings $(\maxit, \delta) = (500, 0.2), (2000, 0.2), (2000, 0.1)$ with about a 2 days budgets. The tuning parameter $\gamma$ is set to 10000.
The behavior of planning with the learned CoverLib is visualized in Fig.\;\ref{fig:pr2_visualization}.
We compared our method's performance against several baselines: Global, NNLib(full), and NNLib(selected). Unlike Domains 1 and 2, this domain does not involve differential or equality constraints, allowing us to test various global planners using off-the-shelf well-tuned implementations in OMPL. The selected planners include RRTConnect, Bidirectional EST (BiEST) \cite{hsu1997path}, Bidirectional KPIECE (BKPIECE), Lazy BKPIECE (LBKPIECE) \cite{sucan2011sampling}, Batch Informed Trees (BIT*) \cite{gammell2015batch}, and Adaptive Informed Trees (AIT*) \cite{strub2022adaptively}. For both EST and KPIECE families, we opted for bidirectional versions, as our testing revealed them to be more than an order of magnitude faster than their unidirectional counterparts in this domain. Note that while KPIECE was originally proposed as a unidirectional planner for kinodynamic planning in \cite{csucan2009kinodynamic}, its extension to kinematic planning has proven highly effective, particularly in its bidirectional implementation \cite{sucan2011sampling}. Also, although BIT* and AIT* are asymptotically optimal planners, it is shown that they can sometimes be faster than RRTConnect in finding feasible paths, letting us to include them in our comparison. In the benchmark, BIT* and AIT* are stopped immediately after finding a feasible path. Our choice of planners was motivated by the benchmark experiments presented in the comprehensive comparative review \cite{orthey2023sampling} of modern sampling-based planners. All implementations were sourced from OMPL and used with default settings. The results for PPC and LGC are shown in Fig.\;\ref{fig:result_pr2} and Fig.\;\ref{fig:result_pr22} respectively.

\subsection{Domain 4: Kinematic planning for 8-DOF robot in Cellular Automata generated environment}
\begin{figure}[t]
    \centering
\includegraphics[width=1.0\linewidth]{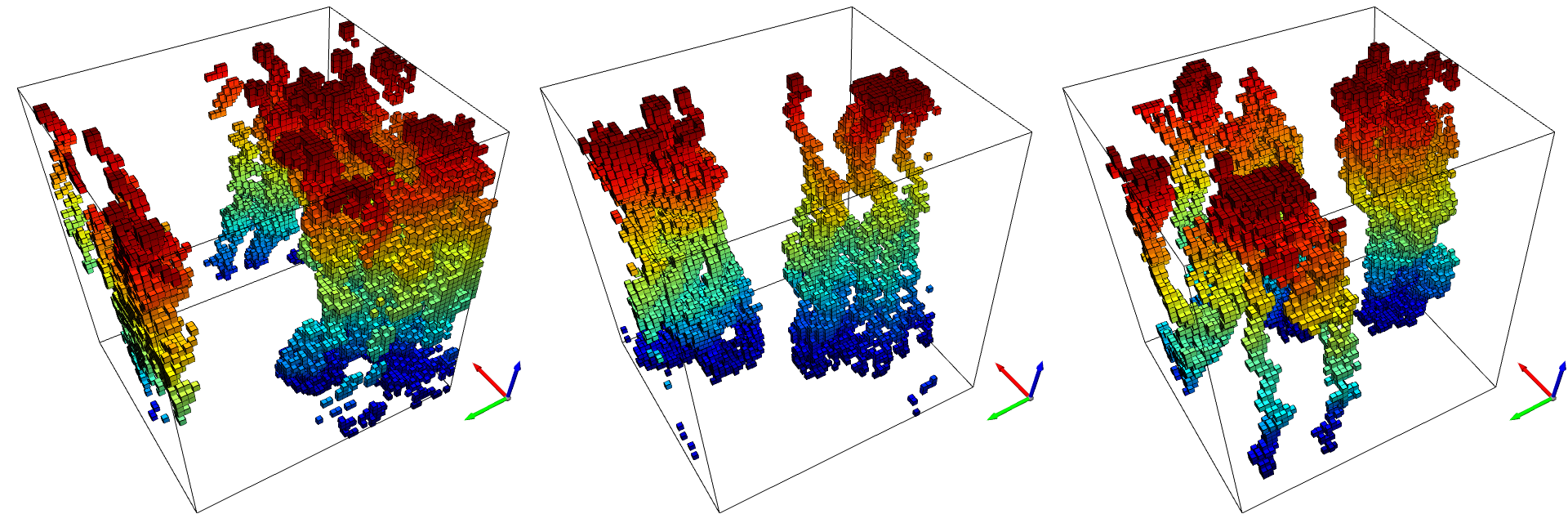}
    \caption{Example of randomly sampled obstacles inside the container in Domain 4. The obstacles are generated by applying Conway's Game of Life rules.}
    \label{fig:conway_jail}
\vspace{-4mm}
\end{figure}

\textbf{Domain definition ($n_{\mathrm{p}} = 410$):}
We utilize the 8-DOF mobile manipulator robot Fetch model. The task goal is to reach a target position inside a container while avoiding procedurally generated ``limestone-cave" like obstacles. The generation process for these limestone-cave shaped landscapes follows these steps: We begin by randomly selecting an x-y point from the bottom surface of the container. Then, we establish a $10\times 10$ pixel grid centered at this point and assign binary values to them randomly. Finally, we apply Conway's Game of Life rules vertically from the bottom layers, where each time step corresponds to the container's height. We repeat these steps multiple times independently, with the repetition number chosen randomly from 1 to 4. We take the union of all the generated grids and consider it as the obstacle. Fig.\;\ref{fig:conway_jail} shows sampled obstacles generated using this method. The environmental configuration variations require 408 DOF. Each single cave structure's DOF consists of 2 dimensions for its x-y position and 100 dimensions for its 10$\times$10 pixel grid, totaling $4 \times (2 + 10 \times 10) = 408$ DOF. While the start robot configuration remains fixed, we randomly select the reaching target position (2 DOF) from the y-z plane at the container's far end, resulting in $n_p = 410$. We define the P-space as the Cartesian product of $\mathbb{R}^2$ and $\{0, 1\}^{56\times 56 \times 56}$. This comprises the target position and the binary occupancy grid of size $56 \times 56 \times 56$ for the container. The motion planning problem follows the same definition as Domain 3, with the only difference being that here the self-collision is considered. We compute the SDF based on the distance to the nearest voxel's center, using a KD-tree for efficient computation.

\textbf{Adaptation algorithm} and \textbf{From-scratch planner} remain the same as Domain 3.

\textbf{Cost predictor:}
We adopt the vector-encoder modeling for $\bar{f}$, where the encoder part is a 3D CNN. The input is $56\times 56 \times 56$ binary values. The 3D CNN encoder is pre-trained as described in Section \ref{sec:nn_architecture}. The pre-training time, including dataset generation, was about 313 minutes on \desktop.

\begin{figure}[t]
    \centering
\includegraphics[width=0.95\linewidth]{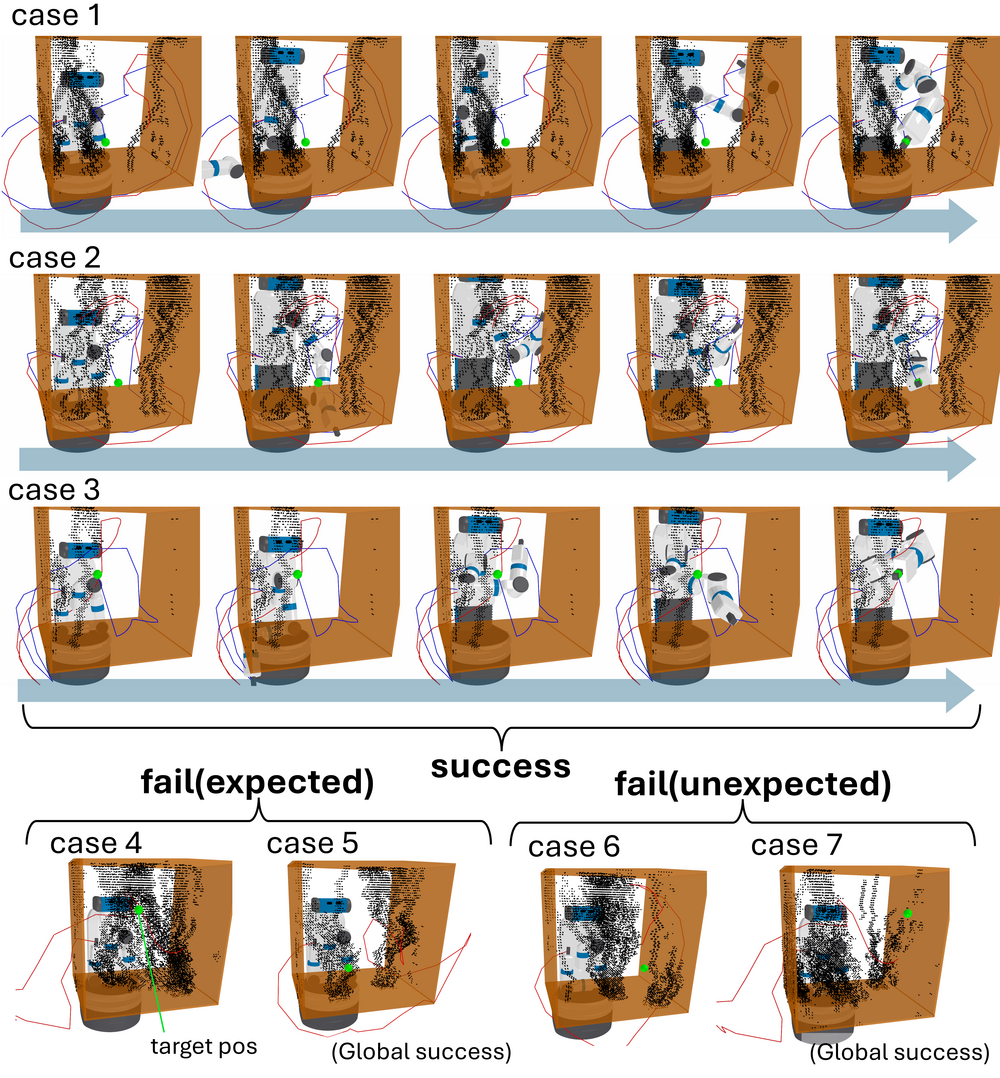}
    \vspace{-5mm}
    \caption{This figure visualizes the behavior of the CoverLib-aided planning with $(\maxit, \delta) = (50000, 0.2)$ in Domain 4. For caption details, refer to Fig.\;\ref{fig:humanoid_visualization}.}
    \label{fig:conway_visualization}
\end{figure}

\begin{figure}[t]
    \centering
    \includegraphics[width=1.0\linewidth]{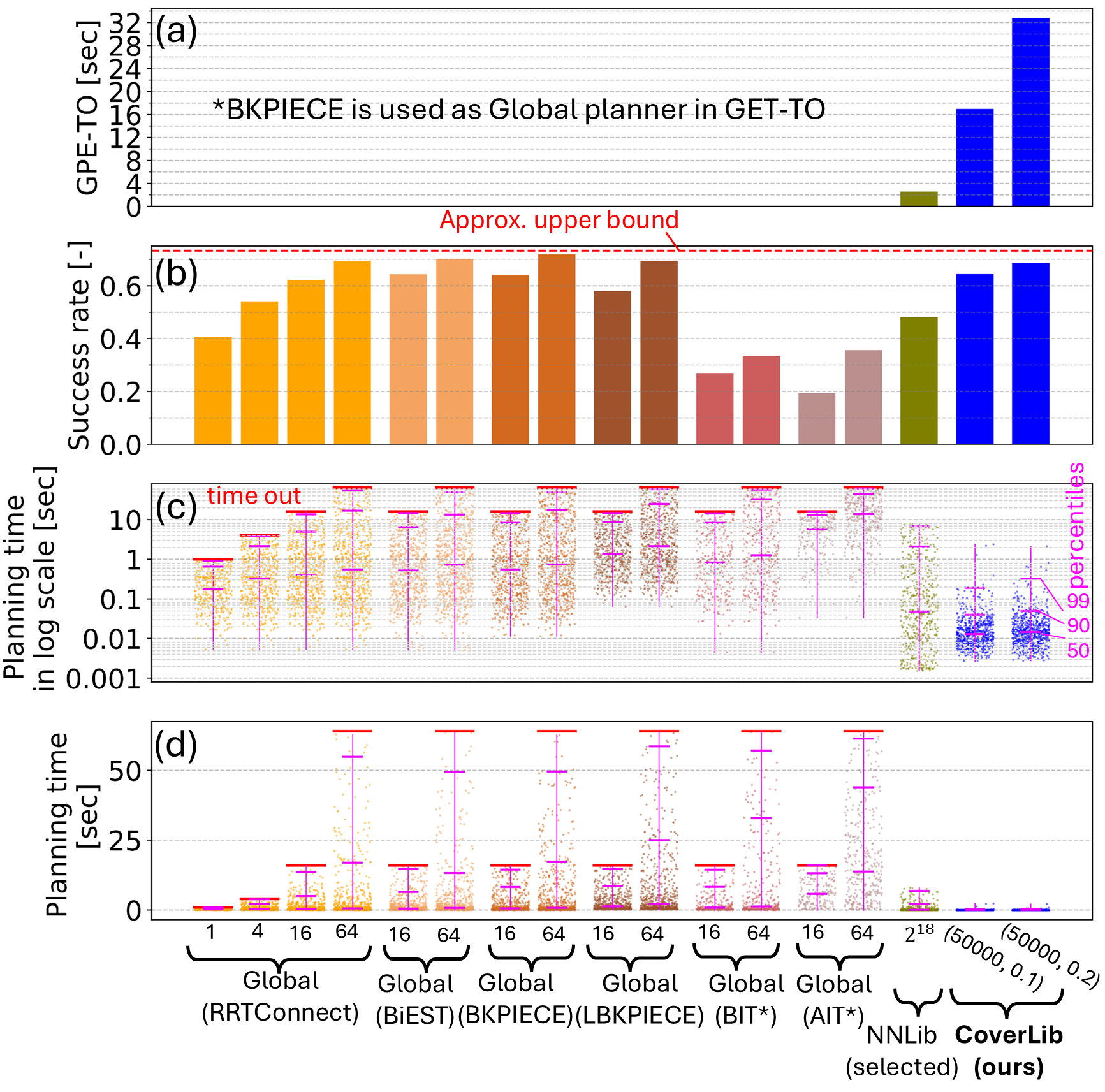}
    \caption{Planning performance comparison for Domain 4. For caption details, refer to Fig.\;\ref{fig:result_double_integrator1}.}
    \label{fig:result_conway}
\end{figure}

\begin{figure}[t]
    \centering
    \includegraphics[width=1.0\linewidth]{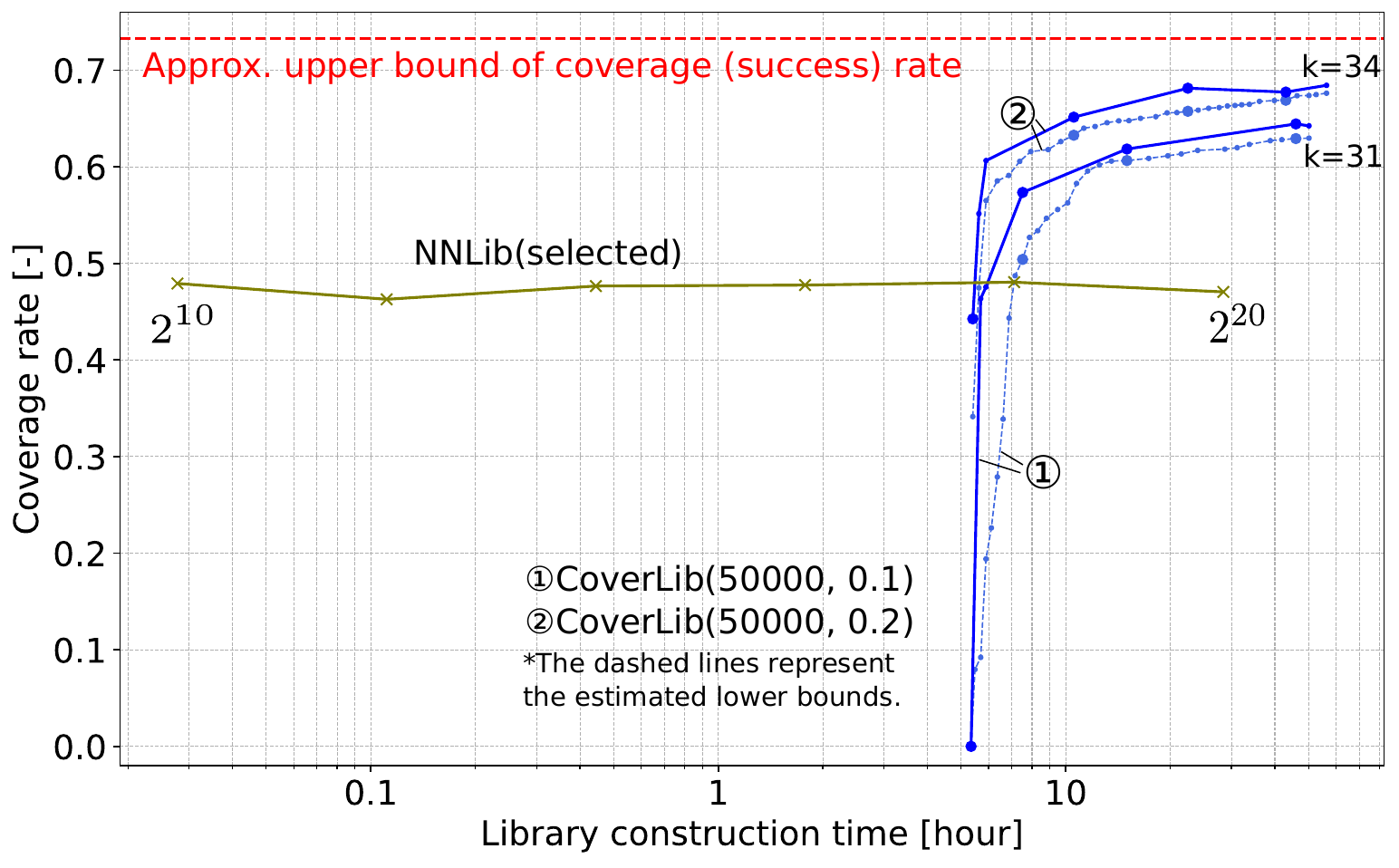}
    \caption{Library growth comparison for Domain 4. For caption details, refer to Fig.\;\ref{fig:result_double_integrator2}.}
    \label{fig:result_conway2}
    \vspace{-5mm}
\end{figure}

\textbf{Result:}
The CoverLibs are constructed with multiple settings $(\maxit, \delta) = (50000, 0.1), (50000, 0.2)$ with 48 hours budgets. The tuning parameter $\gamma$ is set to 10000.
The behavior of CoverLib is visualized in Fig.\;\ref{fig:conway_visualization}. The performance is compared with the baselines Global and NNLib(selected). The selection of the Global planners is the same as in Domain 3. We omit the comparison with NNLib(full) here because the parameters for the environment are binary values, which do not fit well with nearest neighbor search. The results of the PPC and LGC are shown in Fig.\;\ref{fig:result_conway} and Fig.\;\ref{fig:result_conway2} respectively.


%% file: section/discussion.tex
\section{Discussion} \label{sec:discussion}
\begin{figure*}[t]
    \centering
    \includegraphics[width=0.98\linewidth]{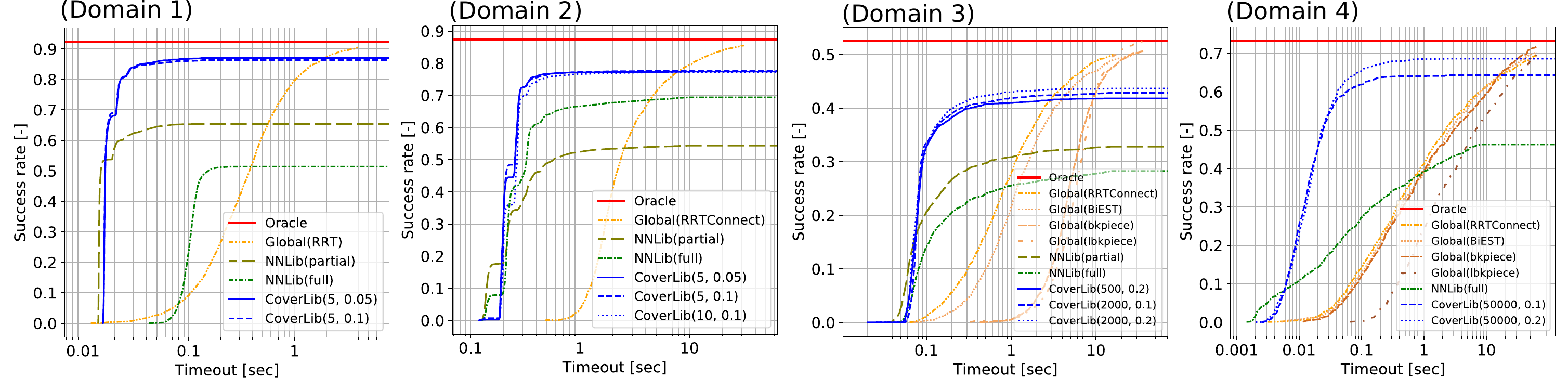}
    \vspace{-3mm}
    \caption{This figure plots the success rate of each planner as a function of the timeout. Note that oracle curve is the upper bound.}
    \label{fig:tradeoff}
    \vspace{-2mm}
\end{figure*}

\subsection{Interpretation of the planning performance comparison (PPC) results}
The PPC results in Figures \ref{fig:result_double_integrator1}, \ref{fig:result_humanoid1}, \ref{fig:result_pr2}, and \ref{fig:result_conway} demonstrate the key advantages of the proposed CoverLib compared to existing approaches. Across all settings, CoverLib achieves higher success rates and substantially higher GPE-TO than NNLibs, indicating that CoverLib-based planning can solve significantly more difficult problems. In terms of planning speed, CoverLib is as fast as or slightly faster than NNLibs in Domains 1 to 3 and is moderately faster than NNLibs in Domain 4, while providing about an order of magnitude speedup over the Global planner. Note that CoverLib's inference overhead for $\bar{f}_{1:K}$ evaluation was 1ms for vector-encoder modeling in Domains 1, 3, and 4, and 0.4 ms for vector modeling in Domain 2. This overhead is visible only in Domain 4's PPC plot (Fig.\;\ref{fig:result_conway}), where CoverLib's minimum planning time lags 1ms behind NNLibs. The NNLibs' nearest neighbor search time is also negligible in most cases. However, for $p(\theta)$ with large $n_p$, NNLib's overhead becomes significant, as shown by NNLib(full) in Fig.\;\ref{fig:result_double_integrator1}, despite using Ball Tree.

To visualize the plannability-speed trade-off between Global and NNLibs and how CoverLib circumvents it, we plot the timeout-success trade-off curves for the Domains 1 to 4 in Fig.\;\ref{fig:tradeoff}. The relationship between Global and NNLibs clearly illustrates the aforementioned trade-off. In contrast, CoverLib significantly circumvents this trade-off, with its curve in closer proximity to the \textit{Oracle} curves. Here the Oracle refers to a hypothetical planner that can solve any feasible problem instantly, thereby achieving the upper bound of the success rate for all timeouts. The upper bound value is computed as described in Section \ref{sec:benchmarking_methodology}. However, the right side of each plot shows that, given a significant timeout, Global solves more difficult problems than CoverLib. Addressing this gap is an obvious avenue for future work and should be pursued by considering the discussion in Section \ref{sec:limitations}.

\begin{figure}[t]
    \centering
    \includegraphics[width=1.0\linewidth]{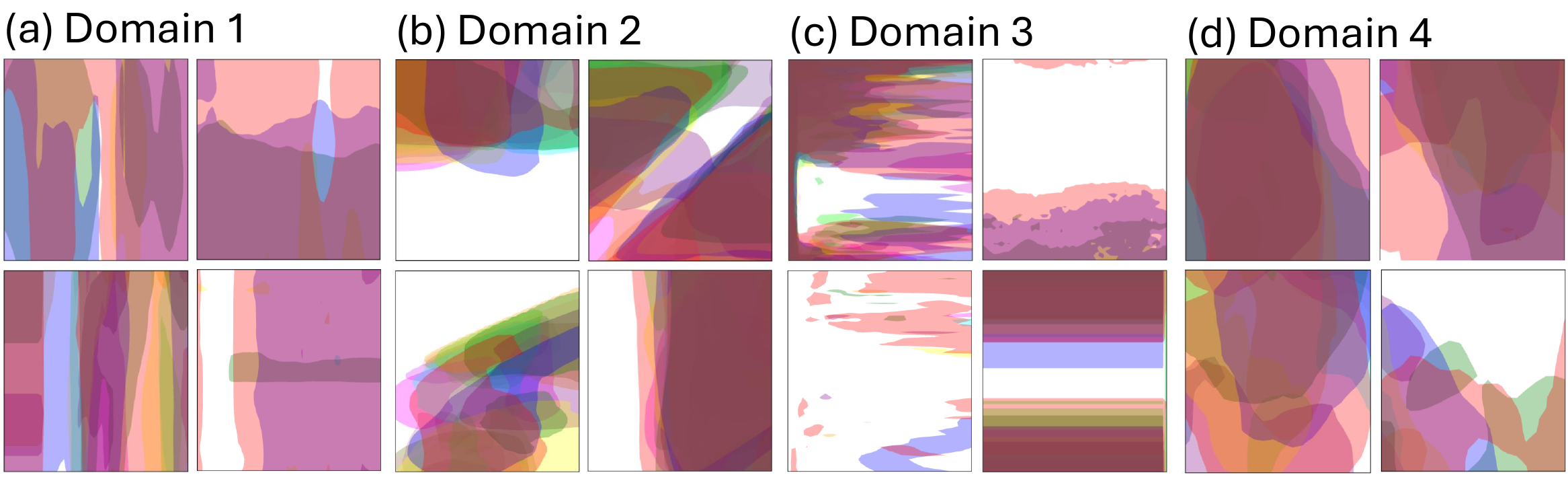}
    \vspace{-5mm}
    \caption{
Visualization of classification results of $\hat{f}_{1:K}$ for Domains 1 to 4. For Domains 1 to 3, among $n_{p}$ dimensions, two axes are selected and values for the other dimensions are randomly chosen and fixed. For Domain 4, among $n_{p}$ dimensions, only those two that account for the target position are used since others are binary values. The lower and upper bounds of the axes are set to the minimum and maximum values of the selected dimensions of 1000 samples from $p(\theta)$. The many classified regions lying throughout the boundary of the axes suggest that the adaptation algorithm has a dimensionality reduction nature, and is captured by Coverlib's classifier. Also, the classified regions often have complex shapes, which are far from those approximated by a sphere or ellipsoid.
    }
    \label{fig:reduction}
    \vspace{-6mm}
\end{figure}

\subsection{Interpretation of library growth comparison (LGC) results}
The LGC plots in Figures \ref{fig:result_double_integrator2}, \ref{fig:result_humanoid2}, \ref{fig:result_pr22} and \ref{fig:result_conway2} show that while NNLibs demonstrates moderate success rates with small library sizes (e.g., low computational time), its success rate growth is quite slow in the end. In contrast, although CoverLib requires dozens of minutes or hours in pre-training the encoder and the first iteration, it quickly catches up with NNLibs and surpasses them in at most several hours, for all four domains. Despite being unable to investigate further due to our computational time budget, the results suggest that CoverLib's success rate continues to grow more actively than NNLibs at the end of the experiments. Note that, the reason for the increase in the time-per-iteration of CoverLib as the iteration goes on is the adaptable determination of $n_{\mathrm{data}}$ explained in Section \ref{sec:step2}.

However, the LGC plots also suggest that NNLibs is more effective than CoverLib if computational resources are limited, as seen at the intersection of the LGC curves between CoverLib and NNLibs. However, we must note that the biggest computational bottleneck of CoverLib can be easily parallelized. Considering the trend of more cores in CPUs and the prevalence of cloud computing, the LGC plots of both CoverLib and NNLibs, and hence the intersection of their curves, will likely shift left in the future. This implies that CoverLib will potentially outperform NNLibs in more scenarios in the future.

NNLib(full) appears to saturate in Domains 1 and 3, which has $p(\theta)$ with large DOF ($n_p=32$ and $n_p=78$, respectively), indicating the impact of the curse of dimensionality. In contrast, NNLib(full) still has room for growth in Domain 2 with a moderate DOF $n_p=10$ of $p(\theta)$. This contrast suggests that NNLib(full) is susceptible to the curse of dimensionality. It is noteworthy that NNLib(selected) consistently outperforms NNLib(full) in the high-dimensional Domains 1 and 3 because feature selection effectively mitigates the curse of dimensionality. However, the observed saturation of NNLib(selected) suggests that the feature selection seems to be less effective in Domain 2. This is potentially because the negative effect of ignoring certain dimensions outweighs the positive effect of reducing the dimensionality.

The LGC plots demonstrate CoverLib's certain robustness to hyperparameters $\maxit$ and $\delta$, as evidenced by similar coverage growth rates across different settings. However, careful tuning of $\delta$ might be crucial when using probabilistic adaptation algorithms. For instance, in Domain 4, where we implement ERTConnect as the adaptation algorithm, the coverage rate increases more slowly with $\delta = 0.1$ compared to $\delta = 0.2$. This behavior stems from ERTConnect's probabilistic nature and its inherent aleatoric error in classification. To maintain a low false positive rate, the algorithm must set the bias very conservatively, which slows down library's coverage growth.

\subsection{Note on memory consumption}
With the library size of $2^{20}$, NNLib consumes a relatively high amount of memory: 1.1GB, 9.1GB, 0.9GB and 4.8GB for the 4 Domains, respectively. In contrast, CoverLib requires only 114MB, 30MB, 63MB and 29MB for the most expensive settings of $(\maxit, \delta)$ in each domain. This significant difference in memory consumption is due to the number of experiences stored. NNLib's individual experiences consume significantly less memory than CoverLib's, as they don't store neural networks. However, NNLib's much larger number of stored experiences outweighs this difference, resulting in higher overall memory consumption.

\subsection{Note on memory and inference time scalability of neural network architecture}
\begin{figure}[t]
    \centering
    \includegraphics[width=1.0\linewidth]{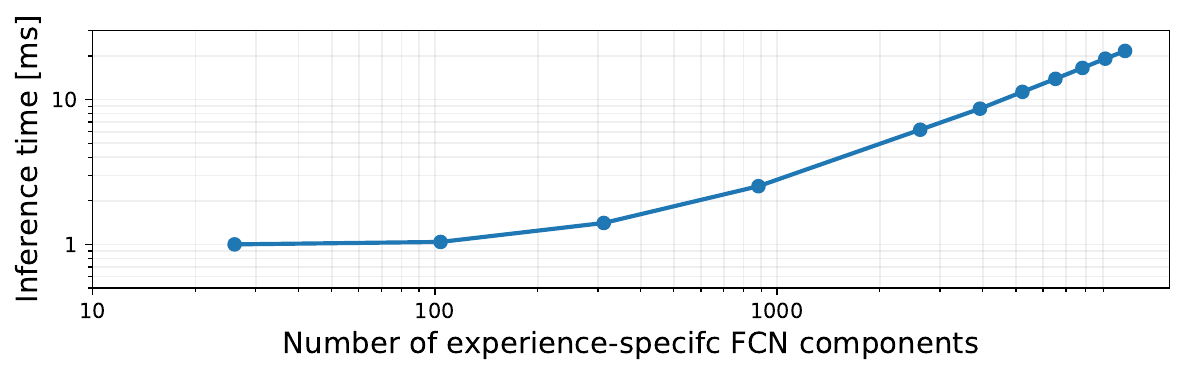}
    \vspace{-8mm}
    \caption{Inference time of the neural network architecture used in Domain 4, with varying numbers of experience-specific FCN components hypothetically.} 
    \vspace{-4mm}
    \label{fig:inference_time}
\end{figure}

A critical consideration for CoverLib is its scalability as domain complexity increases. Complex domains present two key challenges: they require a) more sophisticated neural networks with larger parameters and b) more experiences, each needing its own neural network, potentially leading to memory and inference time constraints.
However, our vector-encoder modeling effectively addresses these challenges. As detailed in Section \ref{sec:nn_architecture}, the architecture employs a shared encoder network across all experiences, requiring only a small, experience-specific component (FCN1 and FCN2 in Fig.\;\ref{fig:vector_matrix_modeling}). This design provides significant memory efficiency. In our setting detailed in the Appendix, each experience-specific component requires merely 0.87MB of memory. To illustrate scalability: with a hypothetical 4GB shared encoder and 10GB GPU memory budget, the system could accommodate approximately 7,000 experiences.
The shared encoder design also accelerates inference speed in two ways. First, the encoder network requires only one evaluation per query, regardless of the library size. Second, as described in the Appendix, FCN evaluations can be efficiently batched due to their identical architectures with different weights. We conducted an experiment to measure inference time with varying numbers of experience-specific FCN components, simulating larger library sizes by copying the FCN components. The results in Fig.\;\ref{fig:inference_time} demonstrate this efficiency, showing that practical inference times are possible even with numerous experiences in the library.

\subsection{Computational cost breakdown of library construction} \label{sec:cost_breakdown}
\begin{figure*}[t]
    \centering
    \includegraphics[width=1.0\linewidth]{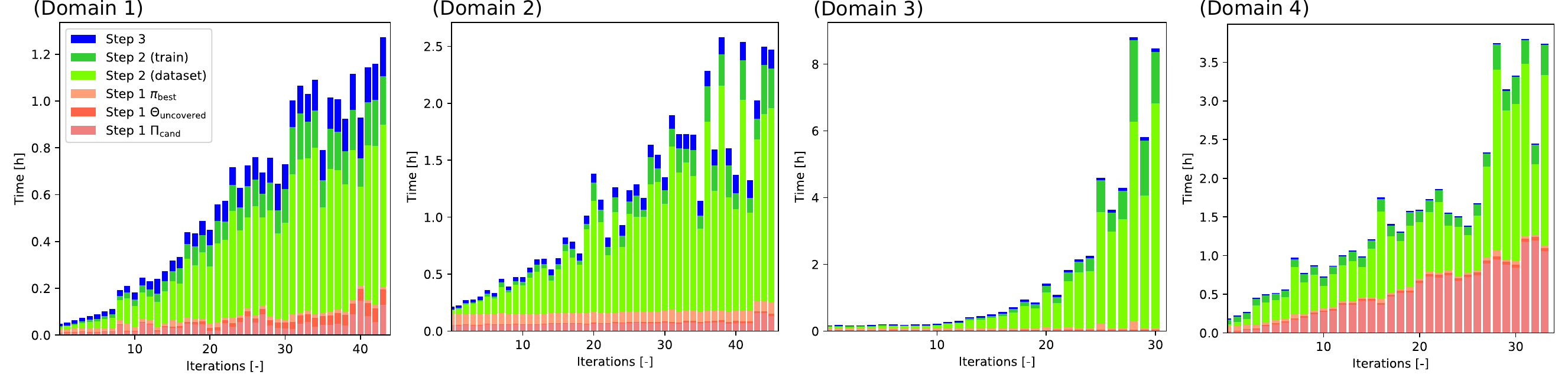}
    \vspace{-7mm}
    \caption{
Time breakdown of CoverLib's library construction process for Domains 1-4, showing the duration of each component across iterations. The stacked bars use six distinct colors to represent the following phases: $\Pi_{\mathrm{cand}}$ sampling, $\Theta_{\mathrm{uncovered}}$ sampling, and $\pi_{\mathrm{best}}$ determination (Step 1); dataset generation and training (Step 2); and bias determination (Step 3).
    }
    \label{fig:breakdown}
    \vspace{-3mm}
\end{figure*}

Log file analysis enables a computational cost breakdown of CoverLib's library construction process as in Fig.\;\ref{fig:breakdown}. Dataset generation (Step 2) dominates the computational time in Domains 1-3, while $\Pi_{\mathrm{cand}}$ sampling as well as the dataset generation are also significant in Domain 4. In Domain 4, the high $\Pi_{\mathrm{cand}}$ sampling time is considered to result from two factors: a relatively high false-positive rate ($\delta=0.2$) and the coverage rate's lower bound nearing the upper bound. Consequently, finding uncovered problems becomes increasingly rare, intensifying the computational cost of finding a solution set $\Pi_{\mathrm{cand}}$ that solves very niche problems. Across all domains, Step 2's computational time increases with each iteration due to the adaptive determination of $n_{\mathrm{data}}$, which is designed to grow progressively as explained in Section \ref{sec:step2}. 

While most domains show linear growth, Domain 3 exhibits exponential behavior in the computational time. This pattern emerges because, according to the log file, Domain 3's achievement rate $\rho_k$ consistently remained below the admissible threshold of $0.3$, triggering repeated multiplication of $\gamma$ by 1.1 in Eq.\;\ref{eq:gamma_update}. The persistently low $\rho_k$ indicates challenges in training an accurate classifier. Two potential factors could explain this: the probabilistic nature of ERTConnect making the classifier's training difficult and inadequacy of using a heightmap representation for an obstacle scene. Given that Domain 4 successfully uses ERTConnect without similar issues, the latter factor is likely to be the primary reason.

\subsection{Ability to capture nonlinear dimensionality reduction nature of the adaptation algorithm and its limitations} \label{sec:limitations}
CoverLib's achievement of larger coverage compared to NNLib, despite using a significantly smaller number of experiences, suggests that the ``small ball assumption" underlying NNLib, discussed in Section \ref{sec:nearest_neighbor}, is invalid. Instead, adaptation algorithms demonstrate an inherent dimensionality reduction property, which the classifiers effectively capture, as shown in Fig.\;\ref{fig:reduction}. The figure shows that the classifier's coverage regions extend throughout the problem distribution in a nonlinear manner, some of the dimensionalities are effectively abstracted away.

A dimension can be abstracted away in two key scenarios: 1) when it has negligible impact on problem solving, and 2) when satisfying constraints in other dimensions facilitates satisfying constraints in this dimension. A dimension in this context refers not to individual axes but to some nonlinear combinations of axes. Also, negligibility here is not binary rather on a continuous spectrum.

\begin{figure}[t]
    \centering
    \includegraphics[width=0.95\linewidth]{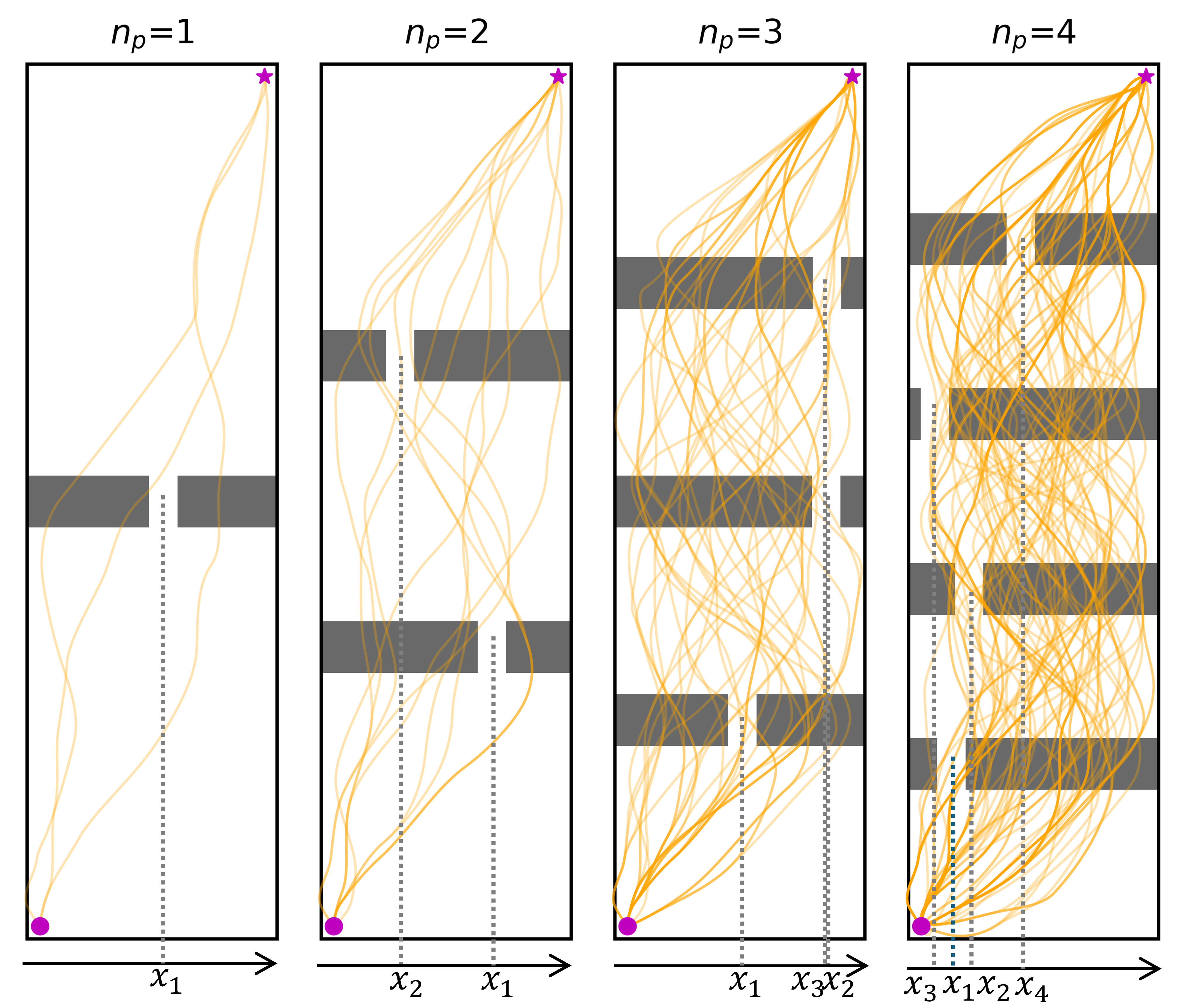}
    \vspace{-3mm}
    \caption{Visualization of the $\mathrm{NarrowPassage}$ domains for different $n_p$ and corresponding libraries created by CoverLib. The x-positions of the narrow passages forms a P-parameter $\theta = (x_1, \ldots, x_{n_p})$.
    }
    \vspace{-3mm}
    \label{fig:parametric_maze_visualize_library}
\end{figure}

\begin{figure}[t]
    \centering
    \includegraphics[width=1.0\linewidth]{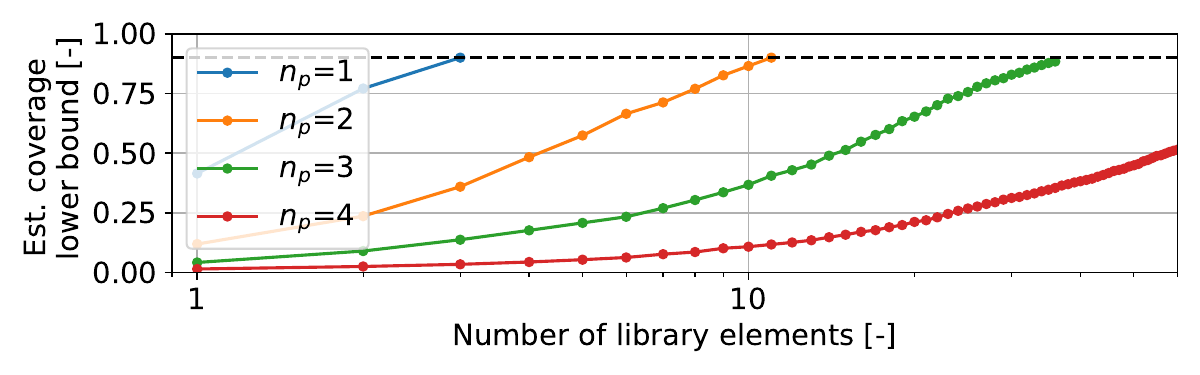}
    \vspace{-8mm}
    \caption{
Coverage rate lower bound estimated by CoverLib versus the number of iterations (library size) for the $n_p$-$\mathrm{NarrowPassages}$ domain class. Since all problems in this domain are feasible, the maximum coverage rate is 1. However, because we set $\delta=0.1$, the estimated lower bound of the coverage rate is at most 0.9.
}
    \label{fig:parametric_maze_result}
\end{figure}

Nevertheless, these observations reveal a fundamental limitation of our method (and library-based approaches in general) when neither condition is satisfied. We demonstrate this through the $n_p$-$\mathrm{NarrowPassages}$ domain class, where $p(\theta)$ has DOF $n_p$ ranging from 1 to 4 (see Fig.\;\ref{fig:parametric_maze_visualize_library}). The domain involves 2D planning between fixed start and goal states, configured similarly to Domain 1. It features multiple same-width narrow passages, with x-positions parameterized by $n_p$ parameters and walls at regular y-intervals. Our numerical experiments show that the required library size grows multiplicatively with $n_p$ (Fig.\;\ref{fig:parametric_maze_result}). This demonstrates that library-based methods are unsuitable for intrinsically complex domains where parameters affect the planning difficulty independently, in a way that satisfying a constraint enforced by a parameter (here, passing a single narrow passage) does not facilitate satisfying constraints imposed by other parameters. This limitation is analogous to data compression algorithms like ZIP or PNG, which excel at compressing structured data but fail with random data.

\subsection{Note on domain shift and generalization} \label{sec:domain_shift}
\begin{figure}[t]
    \centering
\includegraphics[width=0.95\linewidth]{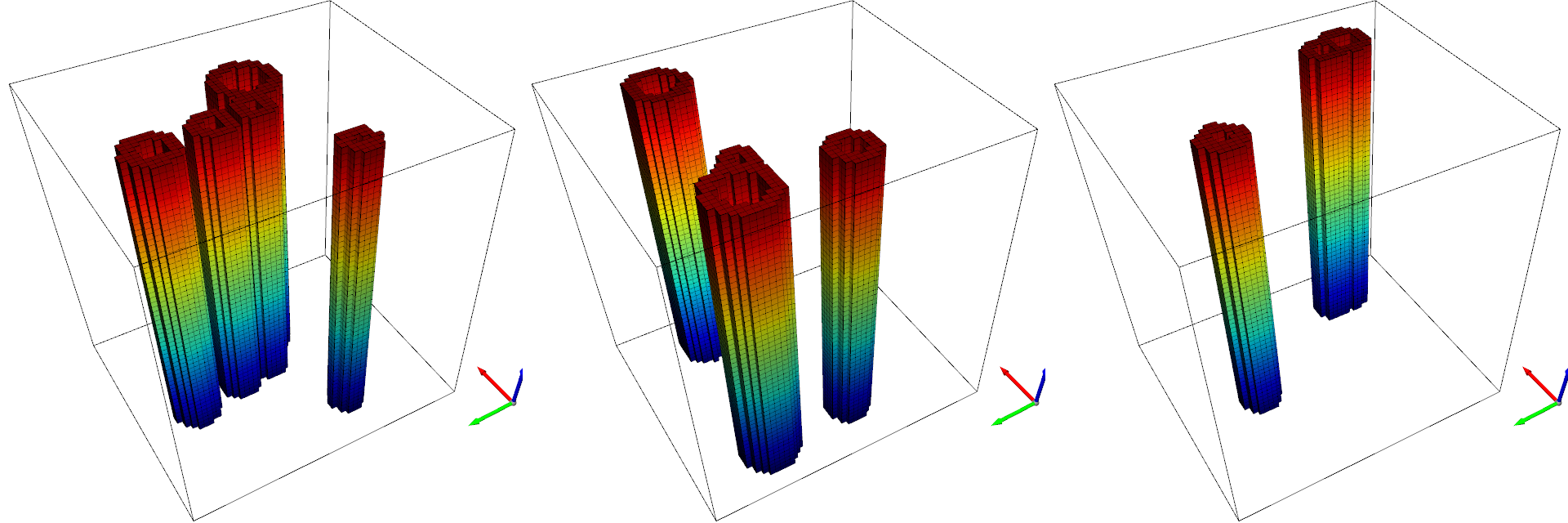}
    \vspace{-2mm}
    \caption{
    Generated multiple vertical bars to test the robustness of Coverlib against domain shifts. The obstacle landscape is distinctly different from the ones in Fig.\;\ref{fig:conway_jail}.
    }
    \vspace{-5mm}
    \label{fig:conway_jail_domain_gap}
\end{figure}

\begin{figure}[t]
    \centering
    \includegraphics[width=1.0\linewidth]{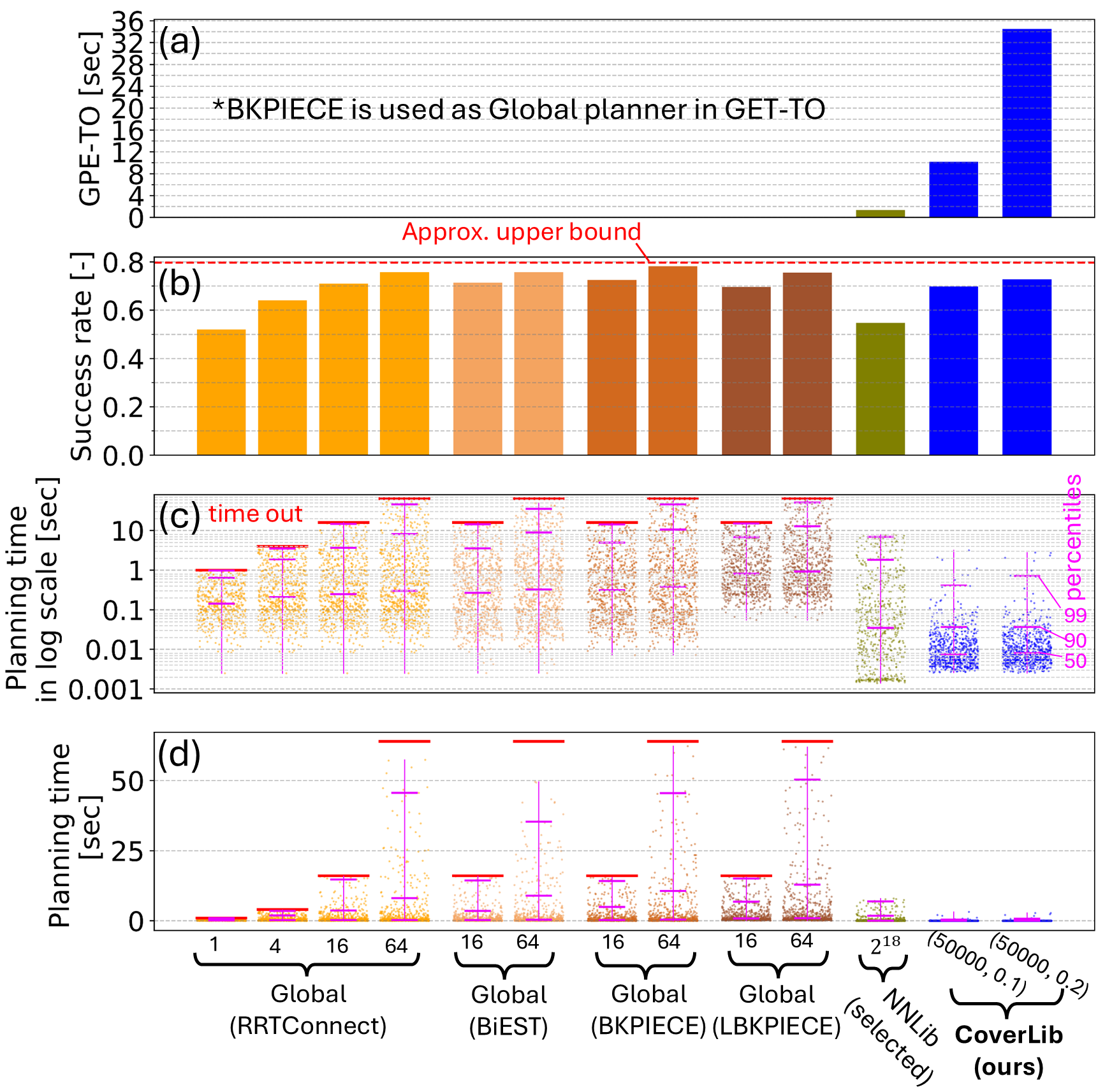}
    \vspace{-5mm}
    \caption{Planning performance comparison for domain-shifted version of Domain 4. For caption details, refer to Fig.\;\ref{fig:result_double_integrator1}.}
    \vspace{-5mm}
    \label{fig:result_domain_gap}
\end{figure}

While beyond the primary scope of this paper's focus on domain-tuned library construction, we observed an interesting result regarding domain shift in a preliminary experiment. We tested a CoverLib trained on Domain 4, without any fine-tuning, on a different $p(\theta)$ featuring multiple vertical bars as Fig.\;\ref{fig:conway_jail_domain_gap} in the container. The geometric pattern is distinctly different from the training distribution's cellular automata-generated structures in Domain 4. Interestingly, even under this distinct domain shift, CoverLib continued to outperform both Global planners and NNLibs as Fig.\;\ref{fig:result_domain_gap}. This robustness suggests that the learned classifiers may be capturing more fundamental properties about adaptability rather than just surface-level environmental features.

\subsection{When to use CoverLib and NNLib, or other learning-based methods} \label{sec:when_to_use}
CoverLib excels when computational resources are plentiful, while NNLib remains effective for problems with small $n_p$ values, because nearest neighbor search is effective in smaller dimensions. In Domain 2, with $n_p$=10 degrees of freedom, CoverLib's GPE-TO only doubles NNLib(full)'s value (Fig.\;\ref{fig:result_humanoid1}). Furthermore, we can see that only in Domain 2, CoverLib takes more than 10 iterations to surpass NNLib(full) in terms of success rate, as shown in Fig.\;\ref{fig:result_humanoid2}. Also, for very small DOF ($n_p \leq 3$), NNLib likely outperforms CoverLib due to its brute-force preparation of all possible solutions.

We will place a special note on the \textit{learning sampling distribution} (LSD) mentioned in Section \ref{sec:contribution}, which is a potential alternative to library-based approaches with ERTConnect. The key distinction is that ERTConnect constrains the search space around a single retrieved experience, while LSD generates sampling distributions based on the collective experiences.
Although it is reasonable to expect that ERTConnect produces a more focused search and thus achieves faster planning speed, the LSD methods offer distinct advantages. LSD's capability to merge collective experiences for generating sampling distributions is valuable when no single past planning experience closely matches the current scenario. This capability makes LSD particularly effective in domains where parameters independently influence planning difficulty, as detailed in Section \ref{sec:limitations}. Furthermore, Chamzas et al. \cite{chamzas2019using, chamzas2021learning, chamzas2022learning} demonstrate LSD's effectiveness in handling completely unseen environmental changes, such as transitioning from shelf environments during training time to table environments during testing time. This kind of change extends beyond the scope of domain shift previously discussed, and library-based methods may not handle it effectively.

%% file: section/conclusion.tex
\section{Future work}\label{sec:future_work}
\textbf{Sample efficient learning of classifiers: } Discussion in Section \ref{sec:cost_breakdown} reveals that dataset generation is the primary computational bottleneck in the library construction process. Reducing this bottleneck requires more sample-efficient classifier training method. Multiple approaches could address this challenge. For example, investigating alternative world representations like point clouds or graphs may provide more efficient alternatives to gridded data. Another approach is enhancing encoder pretraining with motion planning-specific information rather than relying solely on reconstruction loss of the autoencoder.

\textbf{Hardware acceleration for dataset generation: } Leveraging hardware acceleration for both $\Pi_{\mathrm{cand}}$ determination in step 1 and dataset generation in step 2 presents a promising research direction. Both steps are currently bottlenecked by the computational cost of solving a massive number of planning problems. While GPUs are not suitable for solving individual planning problems quickly due to their high latency, their high throughput capabilities make them well-suited for the batch solving of many problems. Given, widespread availability of GPUs, this could significantly improve the framework's accessibility compared to our current experimental setup, which relies on two 80-core \armservers that remain largely inaccessible to many researchers and practitioners.

\textbf{Fight against intrinsically complex domain: }
Our framework handles high-dimensional problem distributions well but struggles with intrinsically complex ones as discussed in \ref{sec:limitations}. A potential solution to this limitation is decomposing the domain into sub-domains and building libraries for each independently. In the $n_p$-$\mathrm{NarrowPassages}$ domain, for example, this means creating subgoals in each ``room" and establishing corresponding subdomain sequences. However, developing such a decomposition methodology in the general case remains an open challenge.

\textbf{Toward real world application: } Our current framework requires users to define a problem distribution a priori at the formulation level. In real-world robotics applications, precisely defining this distribution poses a significant challenge. This challenge inevitably leads to a gap between the defined distribution used for training and the actual distribution encountered during deployment. Our preliminary results in Section \ref{sec:domain_shift} suggest that our framework shows robustness to such discrepancies. Future research should focus on systematically analyzing this robustness and developing methods to enhance it.

\section{Conclusion}
This article addresses the plannability-speed trade-off in motion planning methods. Library-based methods offer a promising approach to mitigate this trade-off when the domain of the planning problem is known a priori. We proposed CoverLib, a principled approach that tunes the \textit{high-level} part of library-based methods to the domain, further alleviating the trade-off.

Numerical experiments demonstrated that CoverLib achieves high plannability (success rate) nearly on par with global planners with high timeouts, outperforming NNLibs in all domains. Simultaneously, CoverLib's planning speed surpassed global planners by more than an order of magnitude by inheriting the speed of library-based methods.
The library growth comparison against NNLibs suggested CoverLib's good scalability with high-dimensional problem spaces. Results indicated that NNLibs are prone to the curse of dimensionality of the DOF $n_p$ of the problem distribution, while CoverLib is less affected by it.
Importantly, the CoverLib method is adaptation-algorithm-agnostic, as demonstrated by the successful integration of both NLP- (in Domains 1 and 2) and sampling-based (in Domains 3 and 4) adaptation algorithms.

We hope this framework inspire future research not only in motion planning, but also in the broader planning-related community, particularly in the MPC, TAMP and Reinforcement Learning fields.